\documentclass[runningheads]{llncs}
\usepackage{graphicx}
\usepackage{hyperref}
\usepackage{tikz}
\usepackage{comment}
\usepackage{amsmath,amssymb} 
\usepackage{color}
\newcommand{\etal}{\textit{et al. }}

\begin{document}
\pagestyle{headings}
\mainmatter
\def\ECCVSubNumber{31}  

\title{Transform Domain Pyramidal Dilated Convolution Networks For Restoration of Under Display Camera Images} 

\titlerunning{PDCRN}
%
\author{Hrishikesh P.S.\thanks{The authors contributed equally} \and
Densen Puthussery$^\star$ \and
Melvin Kuriakose \and
Jiji C.V.}

\authorrunning{H.P.S. et al.}
%
\institute{College of Engineering, Trivandrum, India\\
\email{$\{$hrishikeshps@,puthusserydenson@,memelvin@,jijicv@$\}$cet.ac.in}}
\maketitle

\begin{abstract}
Under-display camera (UDC) is a novel technology that can make digital imaging experience in handheld devices seamless by providing large screen-to-body ratio. UDC images are severely degraded owing to their positioning under a display screen. This work addresses the restoration of images degraded as a result of UDC imaging. Two different networks are proposed for the restoration of images taken with two types of UDC technologies. The first method uses a pyramidal dilated convolution within a wavelet decomposed convolutional neural network for pentile-organic LED (P-OLED) based display system. The second method employs pyramidal dilated convolution within a discrete cosine transform based dual domain network to restore images taken using a transparent-organic LED (T-OLED) based UDC system. The first method produced very good quality restored images and was the winning entry in European Conference on Computer Vision (ECCV) 2020 challenge on image restoration for Under-display Camera - Track 2 - P-OLED evaluated based on PSNR and SSIM. The second method scored $4^{th}$ position in Track-1 (T-OLED) of the challenge evaluated based on the same metrics.

\end{abstract}

\section{Introduction}
\label{sec:intro}
\par 
 
 In recent years there has been a growing demand for casual photography and video conferencing using hand-held devices, especially in smartphones. To address such demands various imaging systems were introduced such as pop-up cameras, punch-hole cameras etc. Under-display camera (UDC) is such a digital data acquisition technique or an imaging system that is a by-product of the growing demand and greater hunger for new technologies. UDC is an imaging system in which a display screen is mounted atop the lens of a conventional digital camera. The major attraction of a UDC device is that the highest possible display-to-body ratio can be realized with a bezel-free design. UDC devices can provide a seamless digital experience through perfect gaze tracking in teleconferences. Leading smartphone makers like OPPO and XIAOMI have already developed and integrated under-display camera technology into some of their flagship products. 

\par The quality of a UDC image is poor due to severe degradation resulting from diffraction and low light transmission. Under-display camera image restoration is relatively a new task for the computer vision community since the technology is new. However, it has a high correlation with many image restoration tasks such as image denoising, deblurring, dehazing and low light image enhancement. The degradations in UDC images can be considered as a combination of degradations addressed in such restoration problems. 
\par
\begin{figure}
    \centering
    \includegraphics[scale=0.35]{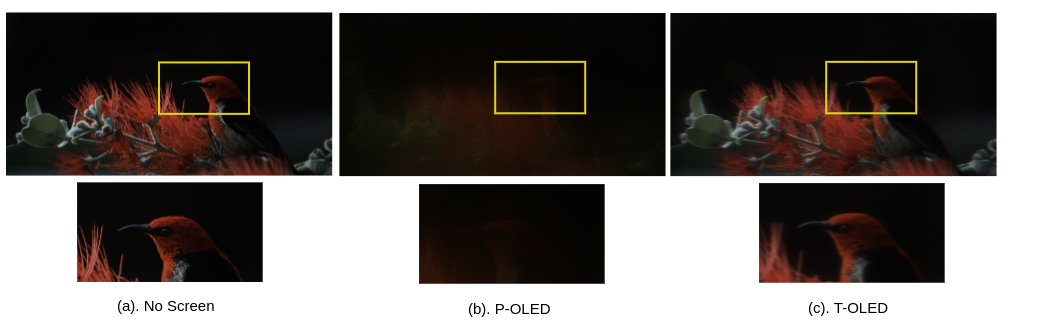}
    \caption{ (a) Original image without display screen (b) degraded image with P-OLED screen mounted on the camera (c) degraded image when using T-OLED screen.}
    \label{fig:degradations}
\end{figure}

 Two different techniques are predominantly used for display purpose in a UDC device. They are transparent organic light emitting diode (T-OLED) and the pentile organic light emitting diode (P-OLED). The type of degradation is different in either of these technologies. The major degradations in T-OLED are blur and noise while in P-OLED they are color shift, low light and noise \cite{udc_paper}. 
Fig.\ref{fig:degradations} shows degraded images taken using P-OLED and T-OLED UDC cameras. The clean image taken without mounting a display screen is also depicted for reference.

\par This work addresses the restoration of under-display camera images taken with P-OLED and T-OLED configurations. Recently, deep convolutional neural network based techniques have proven to be effective for most image restoration tasks. Two methods are proposed in this work to address the restoration of under-display camera images. The first method employs a pyramidal dilated convolution block within a wavelet decomposed convolutional neural network for the restoration of P-OLED degraded images. The second method integrates a discrete cosine transform (DCT) based dual domain sub-net within a fully convolutional encoder-decoder architecture and performs well on T-OLED degraded images.

The first method is a state-of-the-art that won the first position in the European Conference on Computer Vision (ECCV) 2020 challenge on image restoration for Under-display Camera - Track 2 - P-OLED \cite{udc_methods} organized in conjunction with RLQ-TOD workshop. The second method achieved $4^{th}$ position in Track -1 - T-OLED in the same challenge and has comparable performance with the winning methodology in that challenge track.
The main contributions of the proposed work are :
\renewcommand{\labelitemi}{\textbullet}
\begin{itemize}
\item A pyramidal dilated convolutional block has been integrated into a wavelet decomposed encoder-decoder network for state-of-the-art restoration results on P-OLED under-display camera images.
\item An encoder-decoder network with DCT based dual domain sub-net has been developed to address T-OLED under-display camera image restoration.
\end{itemize}
Rest of the paper is organized as follows : in Section \ref{sec:related_wok} we review related works and in Section \ref{sec:proposed_method} we describe the two methodologies that we propose for UDC image restoration. Section \ref{sec:experiments} details our experiments and in Section \ref{sec:result_analysis} we present our result analysis. Finally, Section \ref{sec:conclusions} concludes the work.

\section{Related Work}
\label{sec:related_wok}
\par
Degradation caused by the under-display camera has a significant correlation with other image degradations like real noise, blur and underexposure in images.
Recently, learning based methods have brought a significant improvement in low level vision tasks like  super-resolution \cite{SRGAN,EDSR}, deblurring \cite{vstack},  denoising \cite{denoising_1} etc, and  have produced promising results. A collection of state of the art deep learning techniques on super-resulution and deblurring can be found in \cite{ntire2020_sr} and \cite{ntire2020_deblur} respectively.
\par
Zhou \etal \cite{udc_paper} described the degradation caused by UDC with P-OLED screens as a combination of low-light, color shift and noise. To the best of our knowledge their work was the first intensive study on under-display camera image restoration problem. In their work, they proposed a novel Monitor-Camera Imaging System (MCIS) for paired data acquisition for studying UDC problem. They demonstrated a data synthesizing pipeline for generating UDC images from natural images using only the camera measurements and display pattern. Additionally, they proposed a dual encoder variant of UNet\cite{unet} to restore the under-display camera images they generated using the proposed data pipeline. 
\par
Extensive works have been done in tasks like low-light enhancement and denoising using deep learning techniques. Earlier such tasks were addressed using synthetic data generated using  Additive White Gaussian Noise (AWGN) to develop training set for deep learning based networks. The DnCNN\cite{denoising_paper_zang} network proposed by Zang \etal  performs denoising on noisy images of unknown noise levels. This network follows a modified VGG19\cite{vgg19} architecture.
Recently, there is a shift in the dataset used for training deep learning models. Datasets containing images pairs of original image and images with natural degradation are becoming more relevant. Abdelhamed \etal \cite{denoising_data} proposed novel smartphone image denoising dataset (SIDD) containing images with real noise in smartphone images. Several denoising methods like WNNM \cite{WNNM} and DnCNN \cite{denoising_paper_zang} were benchmarked based on this dataset. Kim \etal \cite{Kim_2020_CVPR} proposed a network that uses adaptive instance normalisation (AIN) to learn a denoiser for real noisy images from the models trained on synthetic training pair. Their network has better generalisation and has produced state-of-the-art results on Darmstadt Noise Dataset (DND) which contains real noisy images.

\par

\par

Liu \etal \cite{mwcnn} proposed a multi-level Wavelet-CNN (MWCNN) for general image restoration tasks. MWCNN follows an encoder-decoder architecture and uses wavelet decomposition for a computationally efficient network for subsampling. In their work, they demonstrated the effectiveness of their method for denoising and super-resolution. The winning team of CVPR NTIRE 2020 Challenge on single image demoireing \cite{ntire2020_demoire} used a wavelet decomposed network similar to MWCNN and achieved state-of-the-art results.

\par Zhang \etal \cite{dmcnn} proposed the use of dilated convolutions for removing compression artifacts from images. In their work, diverse dilation rates were used, with which they reached a receptive field of up to $145 \times 145$. Their theory of receptive field was adopted by Zheng \etal \cite{idcn} wherein they used a pyramidal dilated convolution structure and implicit domain transform to remove compression artifacts. In their work, the dilation rate was gradually increased along with the dilated convolution stack and proved the effectiveness of dilated convolution in removing artifacts with a wide spatial range.

In this work to restore P-OLED degraded images, we propose a fully convolutional network with wavelet decomposition namely  Pyramidal  Dilated  Convolutional RestoreNet (PDCRN) inspired by MWCNN \cite{mwcnn}. Here, the feature extraction in the encoder and decoder levels has been improved by integrating the pyramidal dilated convolution blocks. 
A different approach was followed to address restoration of T-OLED degraded images wherein, a discrete cosine transform (DCT) based dual domain CNN was employed.

\section{Proposed Methods}
\label{sec:proposed_method}
In this section, we describe the two different networks that we propose for the restoration of under-display camera images. Section \ref{sec:proposed_method_poled} details the solution for P-OLED while Section \ref{sec:proposed_method_toled} details the solution for T-OLED images. Although both networks follow a fully convolutional encoder-decoder architecture, the encoder and decoder sub-nets are different in either case.

\subsection{Restoring P-OLED Degraded Images}
\label{sec:proposed_method_poled}
\begin{figure}[h!]
    \centering
    \includegraphics[scale =0.12]{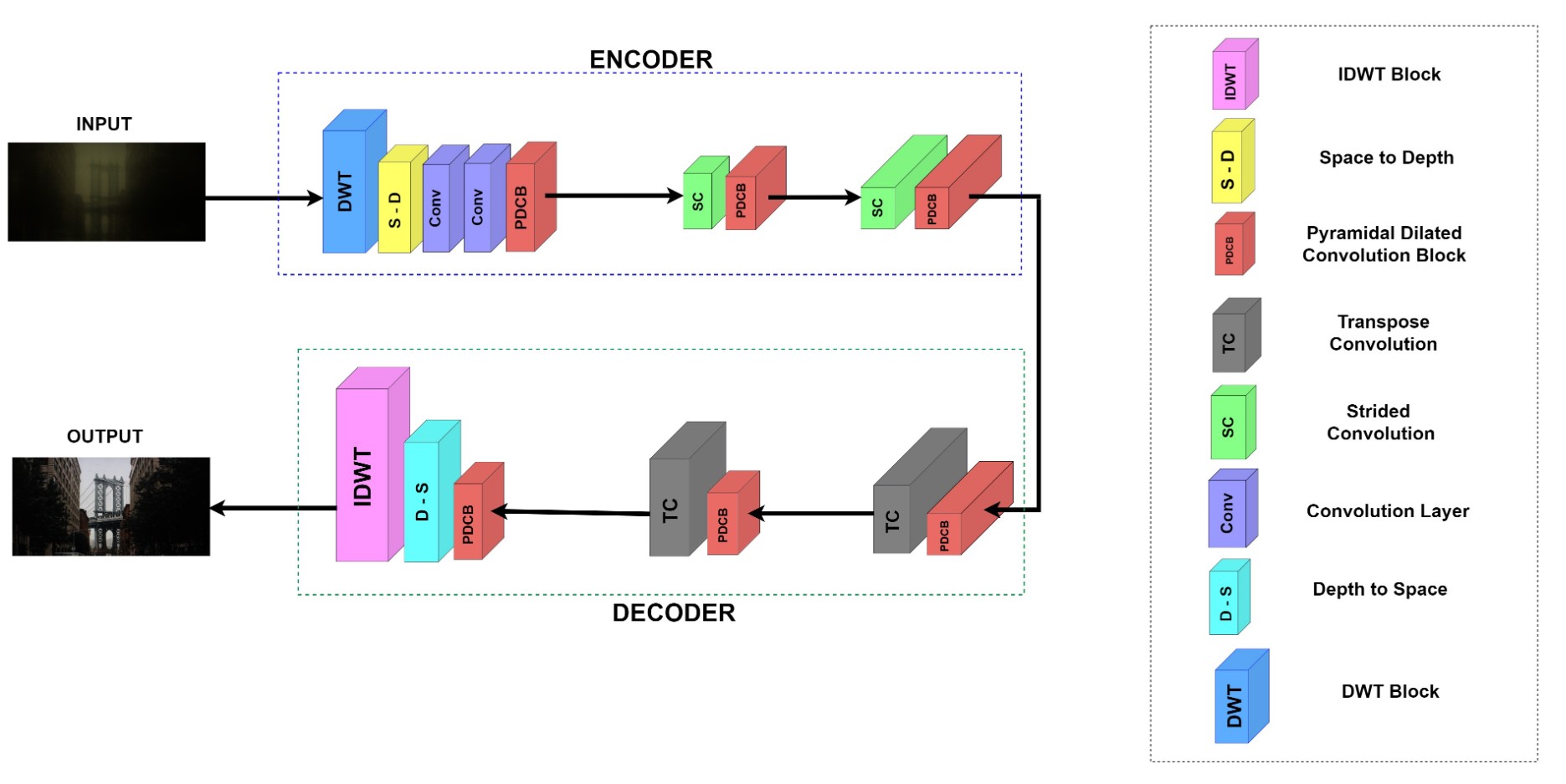}
    \caption{Proposed PDCRN architecture. The network has an encoder and a decoder sub-net as shown in the figure. The network uses PDCB for efficient feature extraction.}
    \label{fig:poled_architecture}
\end{figure}

The proposed Pyramidal Dilated Convolutional RestoreNet (PDCRN) has a fully convolutional encoder-decoder architecture similar to a UNet \cite{unet} as shown in Fig. \ref{fig:poled_architecture}. There are three different operation levels in terms of the spatial resolution of the features being processed. The major degradations in a P-OLED under-display camera image are low light, color shift and noise as mentioned earlier. The task of the encoder is to extract the image features including contrast, semantic information and illumination from the degraded image. The decoder then uses this information to predict the clean image that mimics the one taken without a display screen mounted on the lens. 

\begin{figure}[h!]
    \centering
    \includegraphics[scale =0.08]{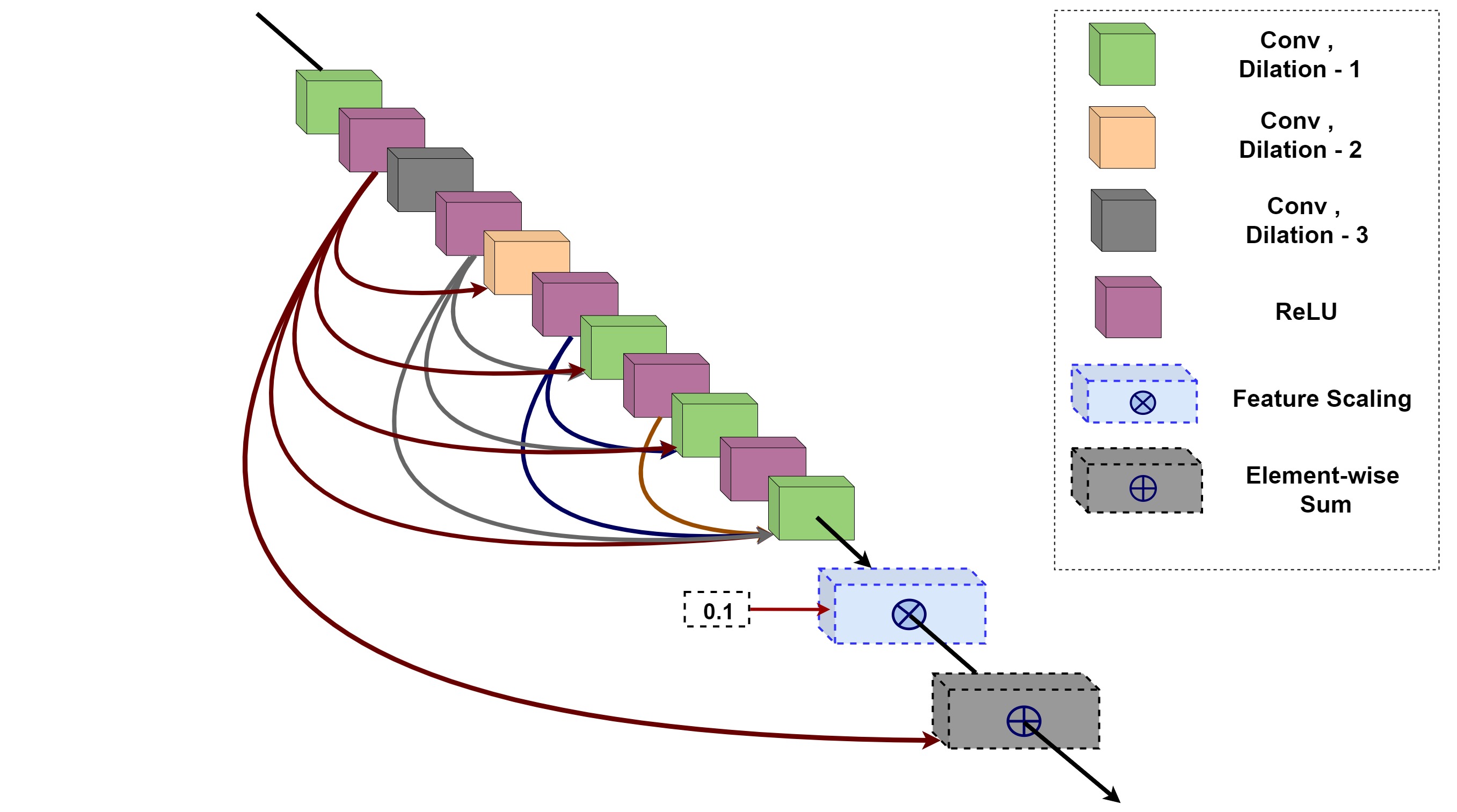}
    \caption{The PDCB sub-net used in PDCRN. The block has a series of densely connected convolutional layers with varying dilation rates for feature extraction.}
    \label{fig:pdcb}
\end{figure}

\subsubsection{PDCRN Encoder}
\par In the encoder section, the input image is processed at $3$ different spatial scales obtained by progressively downsampling the inputs from the previous blocks. At the initial spatial scales of the encoder, the features such as colour distribution, brightness are captured. As encoder progress to smaller spatial scales, higher frequency details like texture, edges etc are learned from the image. The initial downsampling is done using discrete wavelet transform (DWT) based on Haar wavelet as it envelops a larger receptive field which in turn helps in acquiring maximum spatial information from the input image. The downsampling in the subsequent layers is done using strided convolutions as it is empirically found to be more effective than using DWT based downsampling along the whole of encoder. At each level in the encoder, a pyramidal dilated convolution block (PDCB) is used for efficient feature extraction. A detailed description of PDCB is discussed later.

\subsubsection{PDCRN Decoder}
\par The decoder section consists of a series of upsampling operations with PDCB blocks in between to learn the representation of the upscaled image. The decoder progressively upsamples the feature maps from the encoder with $3$ levels of upsampling until the required spatial resolution is reached. The first two upsampling processes are done using transposed convolutions and inverse discrete wavelet transform (IDWT) is used to upsample the image to the original resolution. To make the training process more efficient space-to-depth and depth-to-space\cite{depth_to_space} blocks are used at both ends of the network.
 
\subsubsection{Pyramidal Dilated Convolutional Block (PDCB)}
\label{subsec:pdcb}
\par 
To capture the degradations that have wide spatial extension, convolutional filters of large receptive field is necessary. But this would potentially increase the network complexity and will require a huge dataset to avoid the risk of overfitting. Another solution is to have a sequence of convolutional layers with low filter size which effectively provide a larger receptive field with lower learnable parameters. However, this would also increase the number of trainable parameters and may lead to vanishing gradient problem for deep networks. 
Dilated convolutions with large dilation factors can overcome these limitations. However, large dilation rates may lead to information loss as the aggregated features in two adjacent pixels come from completely non-overlapping input feature set \cite{mwcnn}. Additionally, these may lead to gridding effect in the reconstructed images. To tackle this problem, a pyramidal dilated convolutional block (PDCB) is used for feature extraction in each encoder and decoder levels. 
\par 
PDCB is a residual densely \cite{densenet} connected convolutional network having varying dilatation rates in different convolution layers as shown in Fig.\ref{fig:pdcb}. The block is said to be pyramidal because the dilation rate used in the convolution layers is progressively reduced within the block in which the layers are stacked. Each convolutional layer in the block receives the activation outputs of every preceding convolutional layer as well as the input to the block. At the bottom of the pyramid, the information from far away pixels can be aggregated by using a high dilation rate. The information leakage that occurs with the highly dilated convolution is captured in the top of the pyramid. Thus the network achieves a large receptive field without information loss.

The proposed PDCRN produced excellent results for P-OLED degraded images and the same network was also employed for the restoration of T-OLED degraded images and achieved considerable performance. To further improve the performance for T-OLED images, the network was modified using a dual domain approach with implicit Discrete Cosine Transform (DCT) and obtained superior results for T-OLED compared to PDCRN. This network is explained in detail in the next section.

\subsection{ Restoring T-OLED Degraded Images}
\label{sec:proposed_method_toled}
In a T-OLED display, the size of horizontal  gratings or openings
for light transmission have a size comparable to visible light wavelength \cite{udc_paper}. Due to this, the light is heavily spread in the horizontal axis and results in blurry images. Thus the restoration of T-OLED degraded images is majorly a deblurring problem for horizontal distortion apart from removing the added noise during the acquisition process.

\begin{figure}[h!]
    \centering
    \includegraphics[scale =0.19]{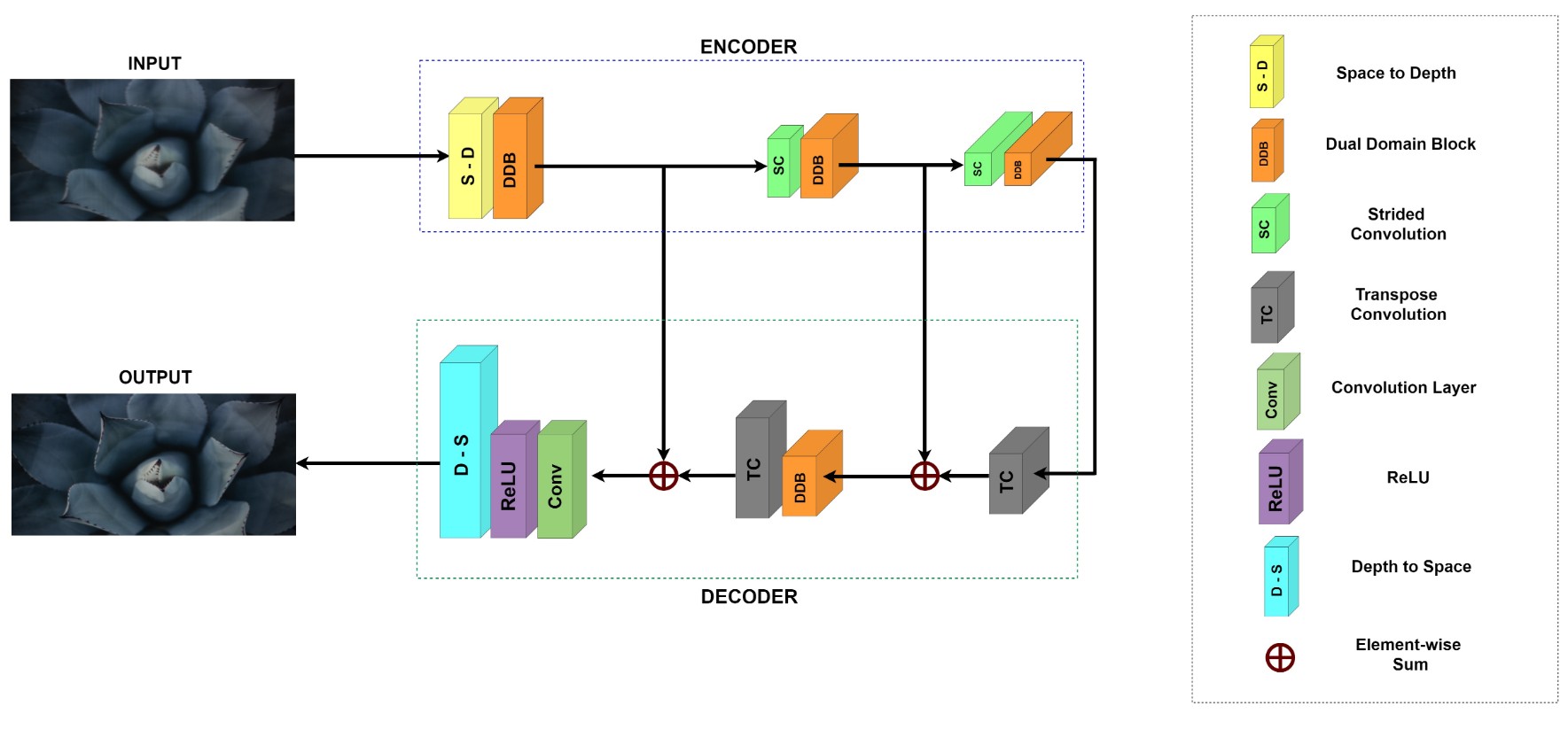}
    \caption{Modified PDCRN architecture with dual domain block (DDB) for T-OLED degraded image restoration. }
    \label{fig:toled_archi}
\end{figure}
The modified PDCRN with dual domain block also follows an encoder-decoder architecture that process the image features at $3$ different scales. The architecture of the modified PDCRN is depicted in Fig. \ref{fig:toled_archi}. Each encoder and decoder level in the network uses a dual domain block with implicit DCT inspired from Implicit Dual-domain Convolutional Network (IDCN) proposed by Zheng \etal \cite{idcn}. IDCN was introduced for restoration of compressed images. This restoration involves correction of details in the high frequency band and removal of distortion artifacts. In a T-OLED degraded image, the restoration task is to remove the horizontally spread blur distortion. since the blur leads to smoothening of edges and other sharp details, the correction is to be done in the mid and high frequency bands. Due to this reason, the IDCN methodology was adopted and improved in this work.

\begin{figure}[h!]
    \centering
    \includegraphics[scale =0.19]{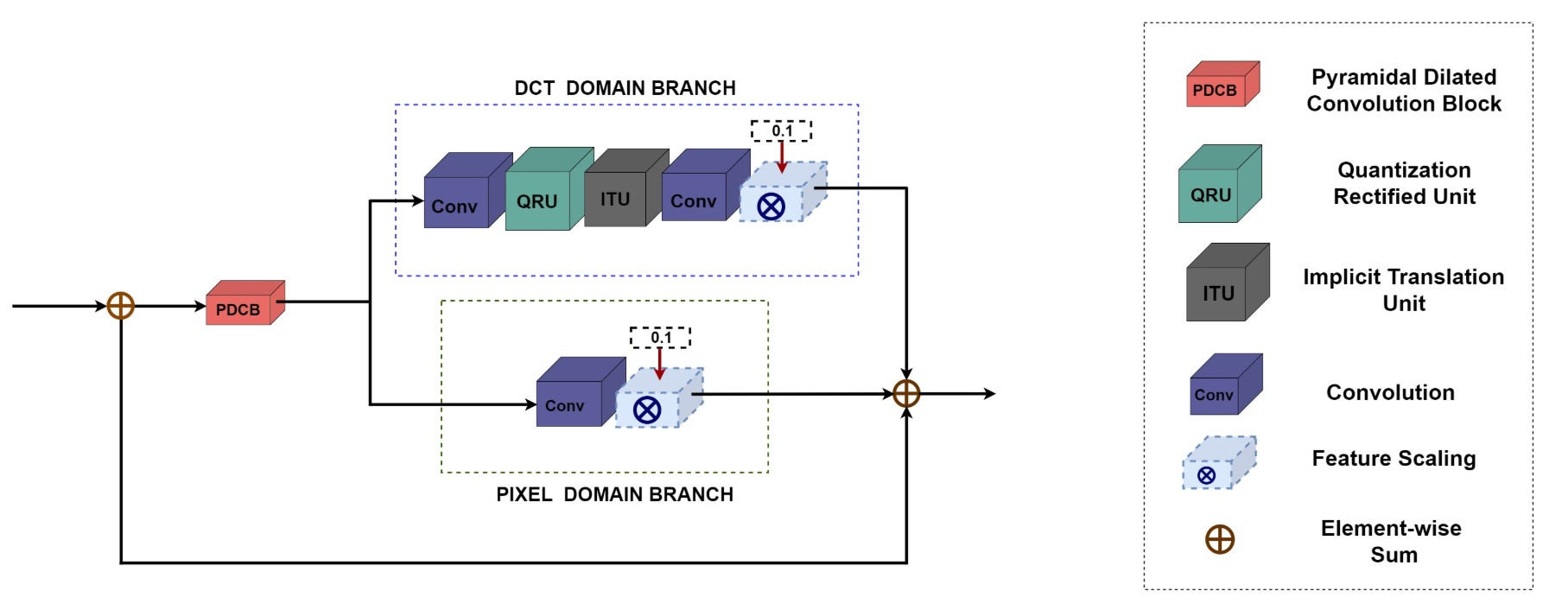}
    \caption{Expansion of the dual domain block in modified PDCRN}
    \label{fig:DDB}
\end{figure}

\subsubsection{Dual Domain Block}
Here we use a simplified version of the IDCN proposed in Zheng \etal \cite{idcn}.
In this block, the image features are processed in the pixel domain using convolution layers and in the frequency domain using implicit discrete cosine transform (DCT).
The DCT domain branch in the dual domain block takes the feature inputs and convert them into the DCT domain using a quantization rectified unit (QRU) and an implicit translation unit (ITU). The QRU thresholds the feature values into a range of $-0.5$ to $+0.5$. This output is given as the input to the ITU where the quantized DCT coefficients of the feature maps are obtained. The DCT domain features are then fused with the pixel domain features obtained from the parallel pixel domain branch as shown in Fig.\ref{fig:DDB}. Thus the dual domain approach performs efficient feature corrections fusing the multiple domain features. The major task of this network is to make corrections in the mid and high frequency details of the image. When DCT is applied to the features from PDCB, the correction in the subsequent layers are happening in frequency domain. This correction will aid the network in learning the edges and other sharp details of the ground truth. 

\subsection{Loss Function}
The network is trained based on mean squared error (MSE) and the MSE loss is the mean squared error between the ground-truth and the predicted images. It is incorporated to generate high fidelity images and is formulated as :
\begin{equation}
    L_{MSE} = \frac{1}{W\times H\times C} \sum_{i=0}^{W-1}\sum_{j=0}^{H-1}\sum_{k=0}^{C-1} (Y_{i,j,k} - \hat{Y}_{i,j,k})^2
    \label{eq:mse_loss}
\end{equation}
where, $W$, $H$ and $C$ are the width, height and number of channels of the output, $Y$ is the ground truth image and $\hat{Y}$ is the predicted image.

\section{Experiments}
\label{sec:experiments}
\subsection{Dataset}
The dataset used in the experiments is the challenge dataset provided by the organizers of RLQ-TOD workshop 2020 for the image restoration for Under-display Camera - Track 1 - T-OLED and Track 2- P-OLED. Both the datasets contains $240$ pairs of degraded under-display camera images and their corresponding ground truth images for training, $30$ pairs for validation and $30$ pairs for testing. Each image is of resolution $1024\times2048\times3$.
\subsection{Training}
\begin{figure}[htbp!]
\centering
\newcommand\x{0.49}
\newcommand\scale{0.35}
  \begin{minipage}{\x\linewidth}
		\begin{center}
		\includegraphics[scale=\scale]{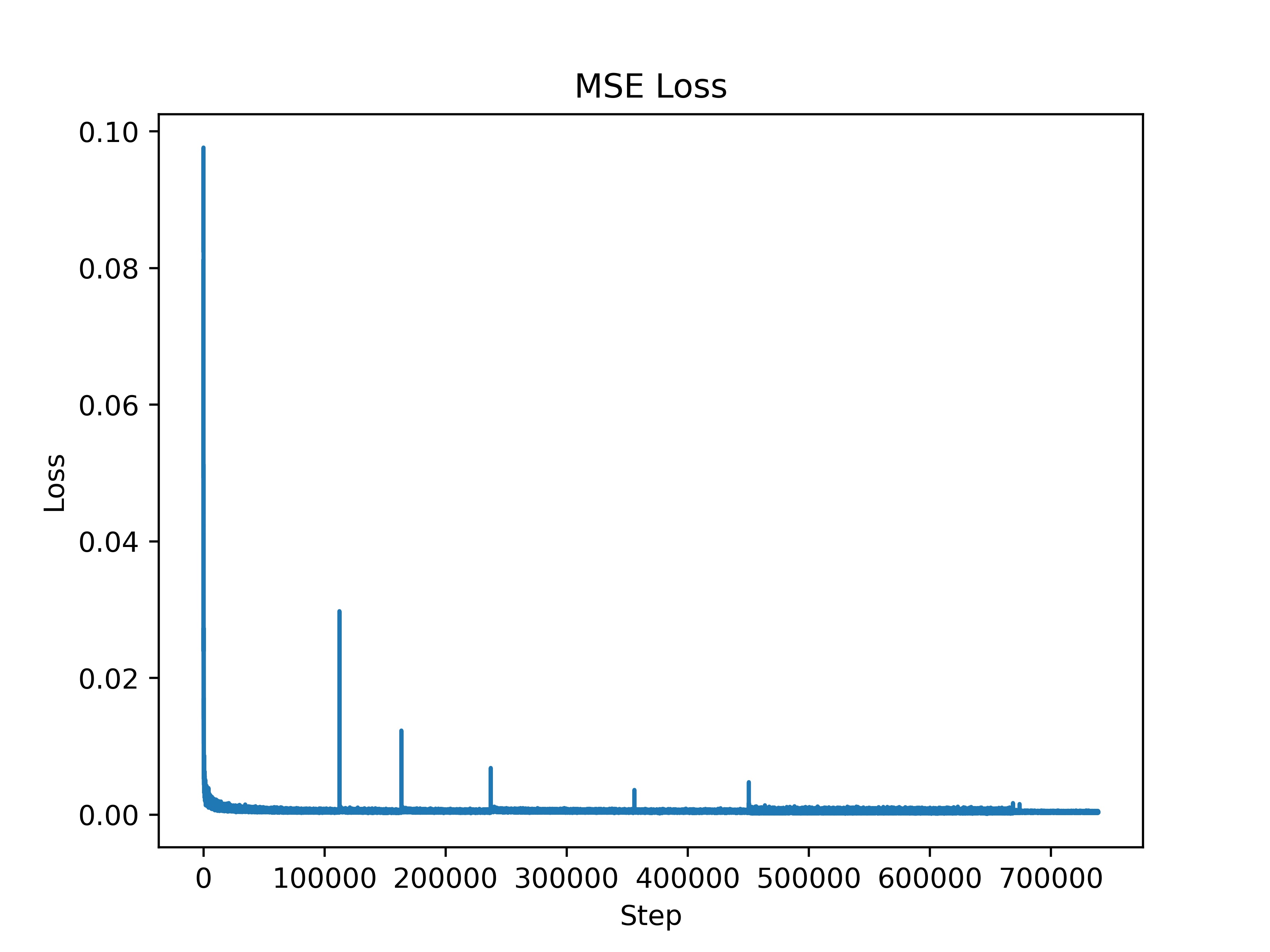}
		\fontsize{8}{12pt}\selectfont (a)
		\end{center}
  \end{minipage}
  \begin{minipage}{\x\linewidth}
		\begin{center}
		\includegraphics[scale=\scale]{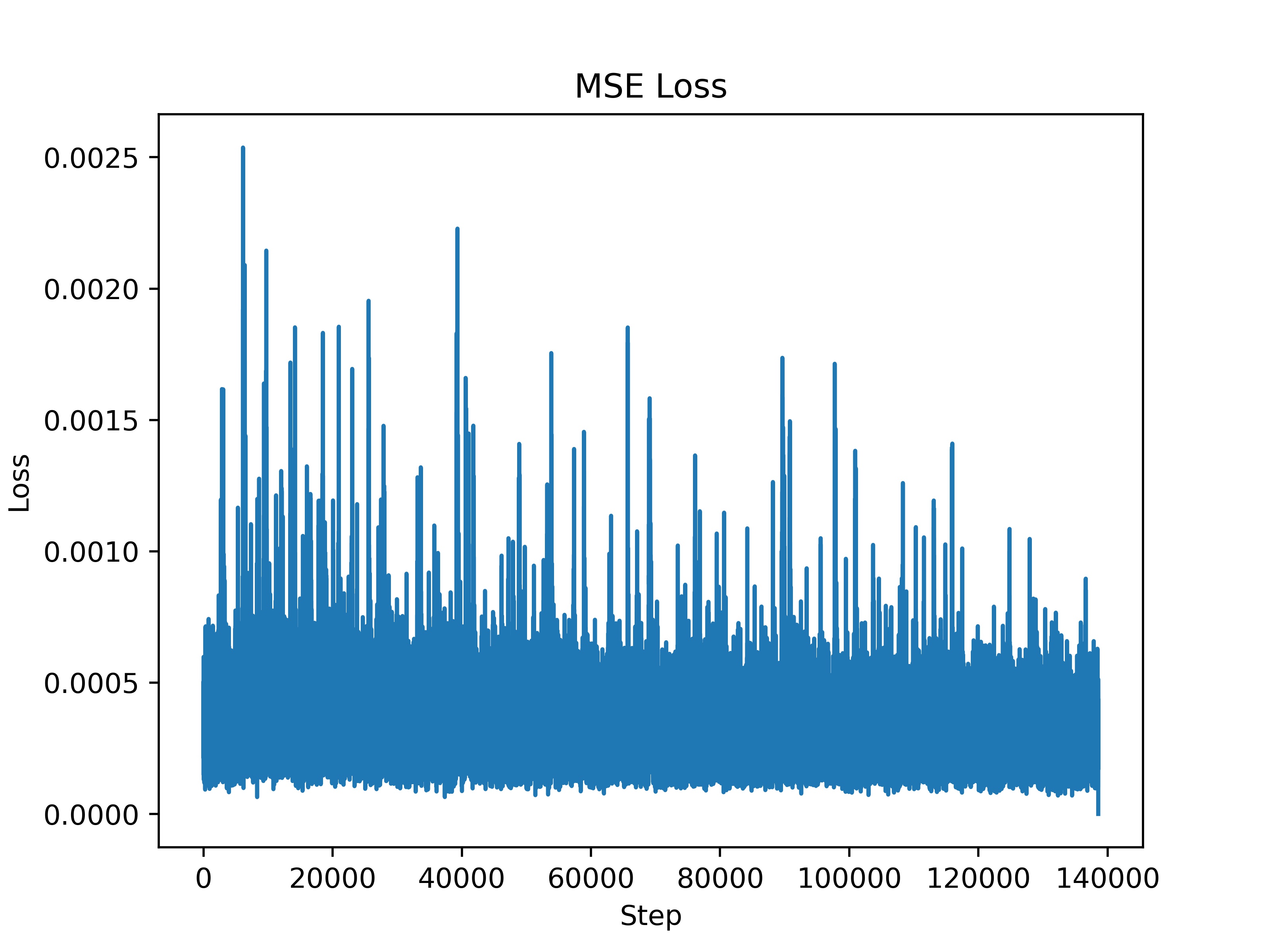}
		\fontsize{8}{12pt}\selectfont (b)
		\end{center}
  \end{minipage}
\caption{Training loss plot of (a) PDCRN for P-OLED (b) PDCRN with dual domain block for T-OLED. The spikes in loss function resulted from discontinuous training.}
\label{fig:loss_plot}
\end{figure}

\begin{figure}[htbp!]
\centering
\newcommand\x{0.49}
\newcommand\scale{0.35}
  \begin{minipage}{\x\linewidth}
		\begin{center}
		\includegraphics[scale=\scale]{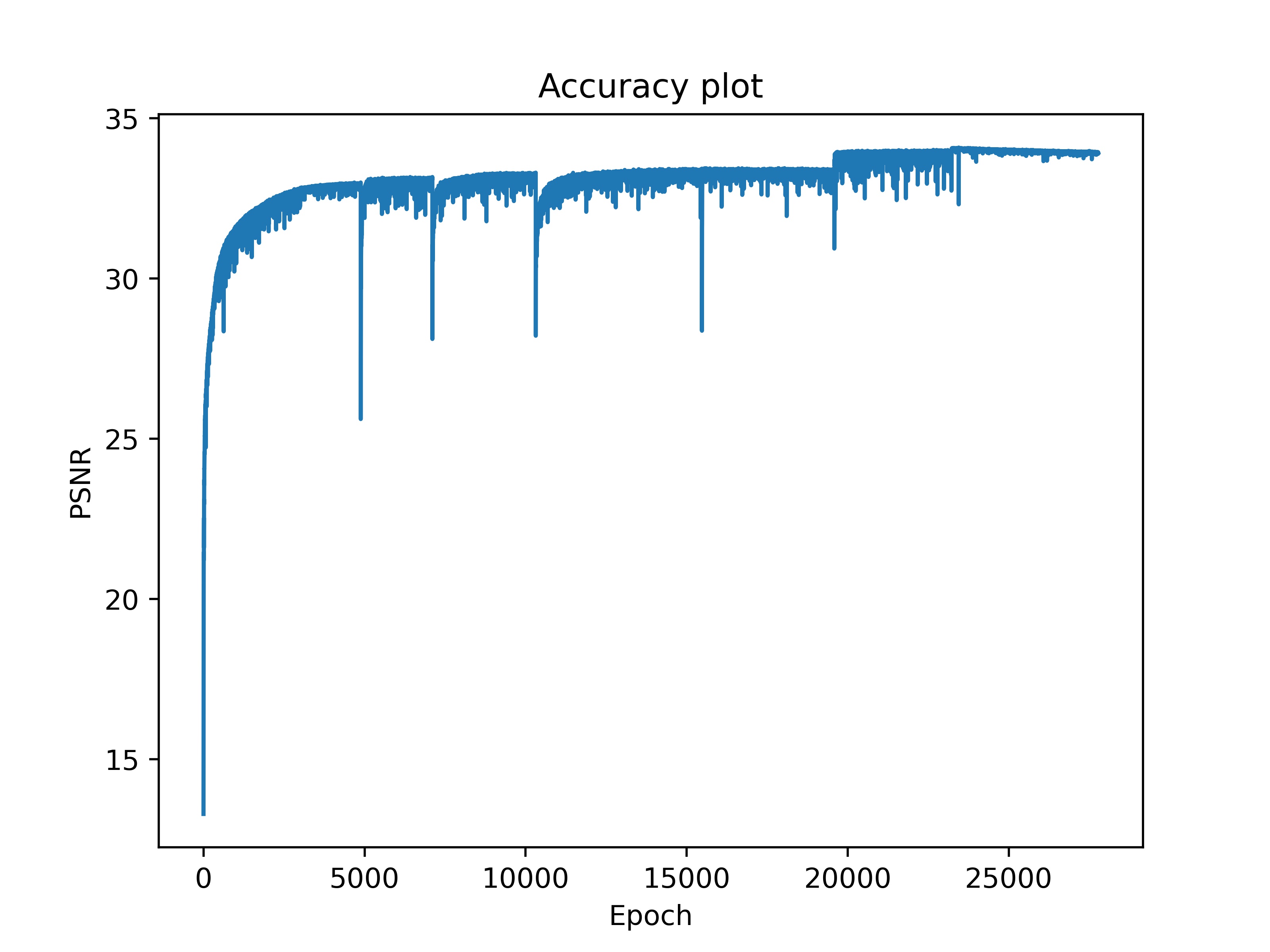}
		\fontsize{8}{12pt}\selectfont (a)
		\end{center}
  \end{minipage}
  \begin{minipage}{\x\linewidth}
		\begin{center}
		\includegraphics[scale=\scale]{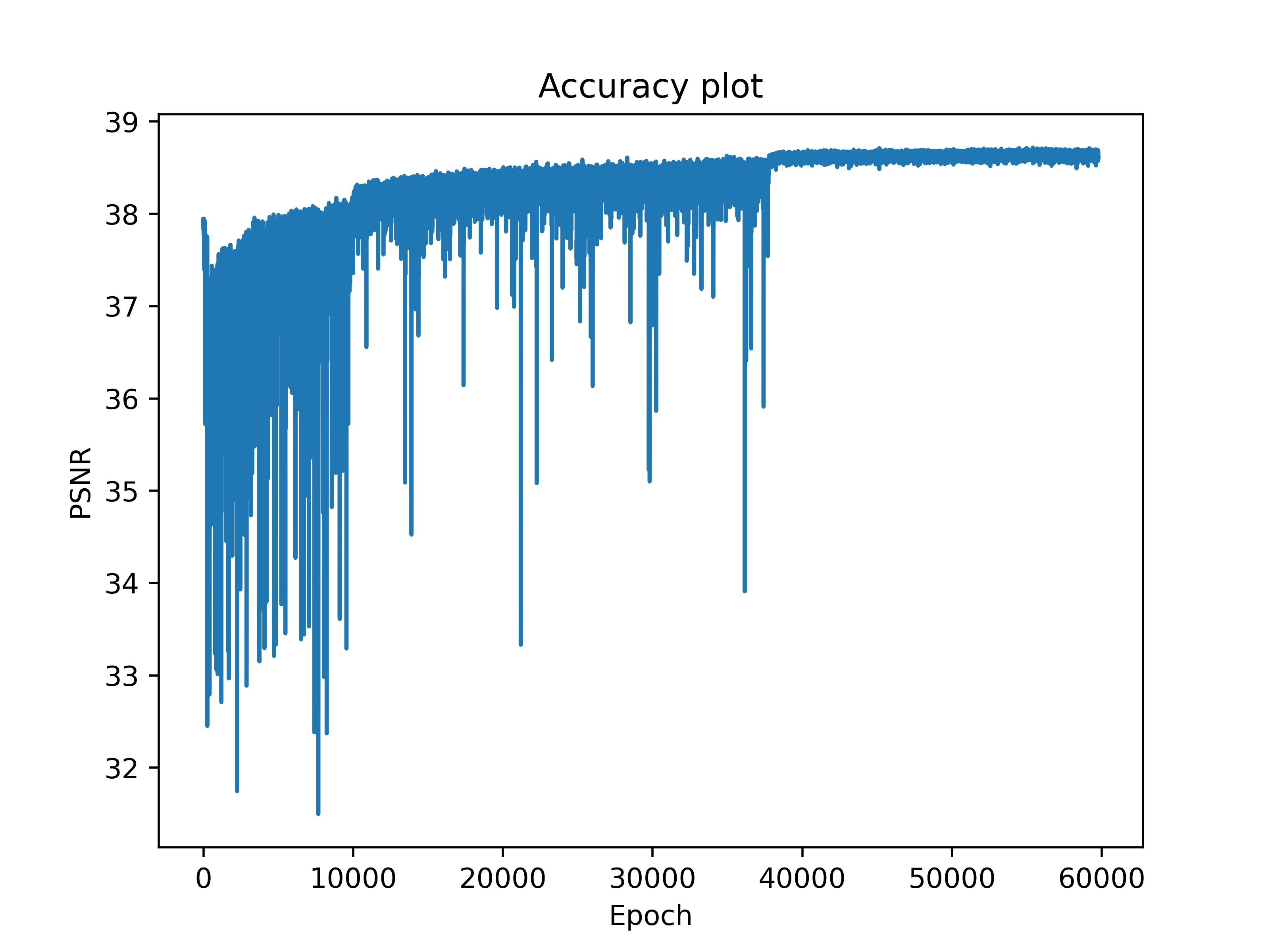}
		\fontsize{8}{12pt}\selectfont (b)
		\end{center}
  \end{minipage}
\caption{Accuracy plot of (a) PDCRN for P-OLED (b) PDCRN with dual domain block for T-OLED. The accuracy plot in (a) is not smooth due to discontinuous training. The accuracy plot in (b) is high in initial epochs because of a high learning rate.}
\label{fig:accuracy_plot}
\end{figure}

 The PDCRN network was trained with images of a patch size of $1024\times2048\times3$ and batch size of $12$. For T-OLED, the images were trained with patch size of $256\times256\times3$ and batch size of $16$.  Adam optimizer with $\beta_1=0.9$ , $\beta_2=0.99$ and learning rate $2e^{-4}$ was used in both the networks. The learning rate is decayed after every $10000$ epoch with a decay rate of $0.5$ in both networks. The networks were trained for a total of $30000$ epochs using Tesla V100 GPU card with 32 Gib memory. The training loss and accuracy plots of PDCRN and dual domain PDCRN are depicted in Fig. \ref{fig:loss_plot} and Fig. \ref{fig:accuracy_plot} respectively.


\subsection{Evaluation Metrics}
To evaluate the performance of the methodology, the evaluation metric used is the peak signal to noise ratio (PSNR) and structural similarity (SSIM).

\subsection{Ablation Study for PDCRN}
\begin{figure}[h!]
\centering
\newcommand\x{0.45}
\newcommand\scale{0.4}
  \begin{minipage}{\x\linewidth}
		\begin{center}
		\includegraphics[scale=\scale]{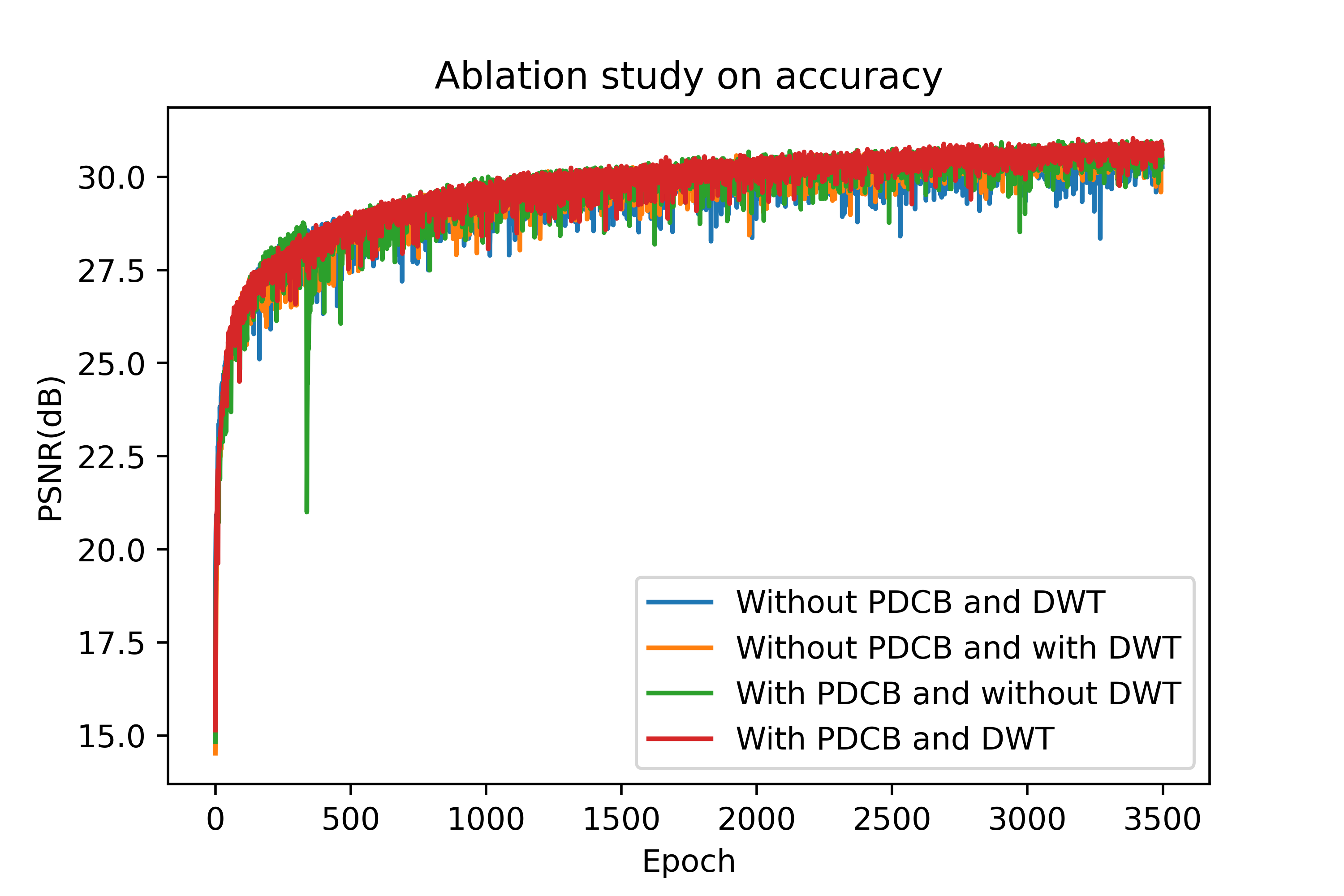}
        \vskip 2pt
        \fontsize{9}{12pt}\selectfont (a) Accuracy plots.
		\end{center}
  \end{minipage}
  \begin{minipage}{\x\linewidth}
		\begin{center}
		\includegraphics[scale=\scale]{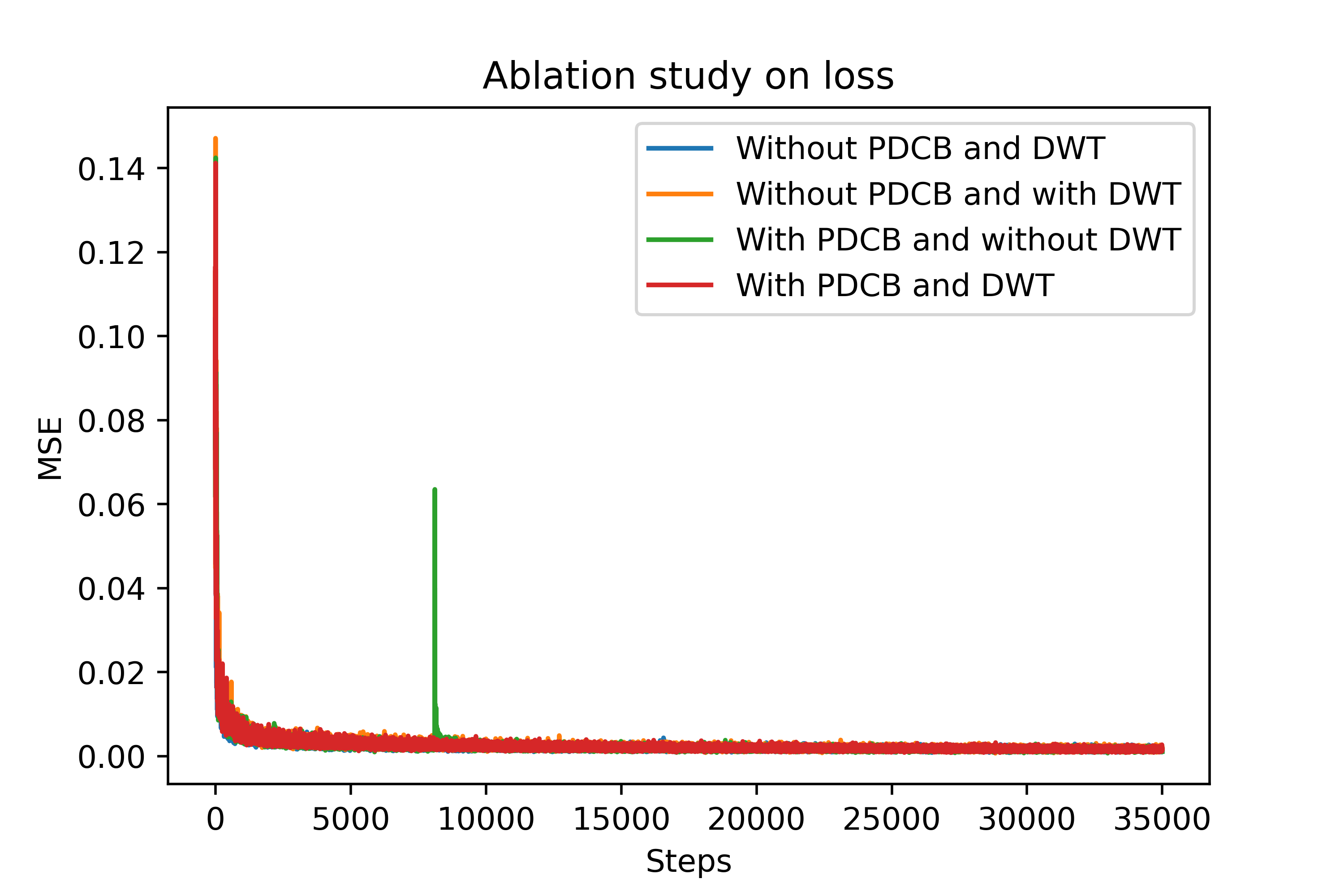}
		\vskip 2pt
		\fontsize{9}{12pt}\selectfont (b) Loss plots.
		\end{center}
  \end{minipage}
\caption{Study of the effects of various components of the proposed PDCRN on accuracy in Fig.(a) and mean squared error(MSE) loss in Fig.(b)}
\label{fig:ablation study}
\end{figure}

The Fig.\ref{fig:ablation study} shows the accuracy and loss curves generated that shows the effectiveness of various components of the proposed network. The accuracy plot is generated based on the PSNR metric in every epoch of training and loss plot is based on MSE loss of every mini-batch of training. For standardisation, all the networks were trained up to $3500$ epochs were the plots reaching a comparative saturation.  From the accuracy plots, we can understand that network without pyramid dilation and DWT performed the worse, attaining a  PSNR of $30.81$ dB. The introduction of dilated convolutions in the pixel domain improved the results to $31 $ dB. The network of the proposed method obtained best PSNR of $30.25$ dB out of all. So with the addition of both DWT and pyramid dilation, there was performance gain of $0.5$ dB when compared to network with a no dilation and DWT. Loss plot doesn't give much insights into the performance comparison. 

\section{Result Analysis}
\label{sec:result_analysis}
\par
\begin{figure}[htbp!]
\centering
\newcommand\x{0.3}
\newcommand\scale{0.050}
  \begin{minipage}{\x\linewidth}
		\begin{center}
		\includegraphics[scale=\scale]{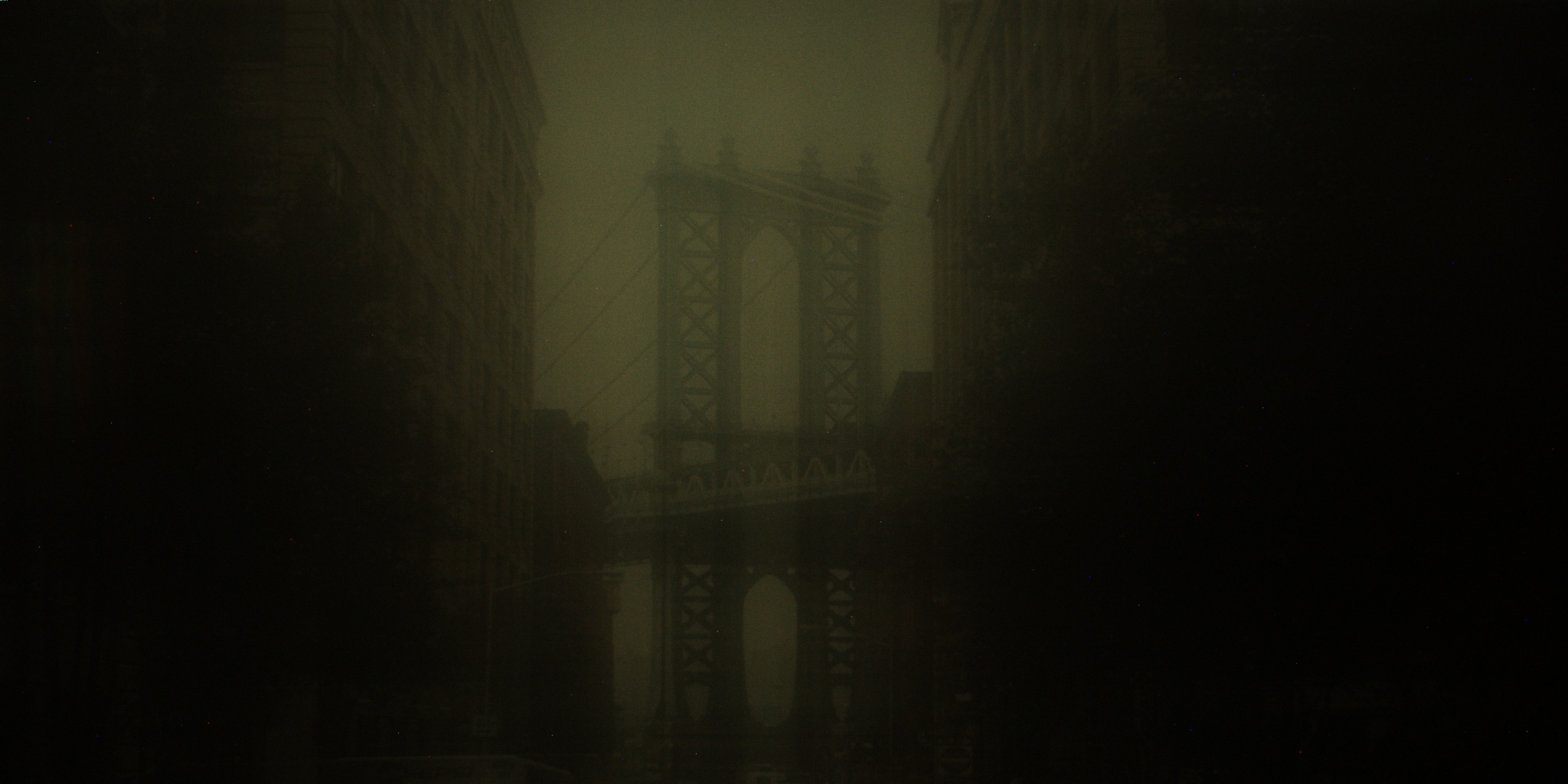}
		\fontsize{8}{12pt}\selectfont PSNR-$13.33dB$/SSIM-$0.63$
		\vskip 2pt
\includegraphics[scale=\scale]{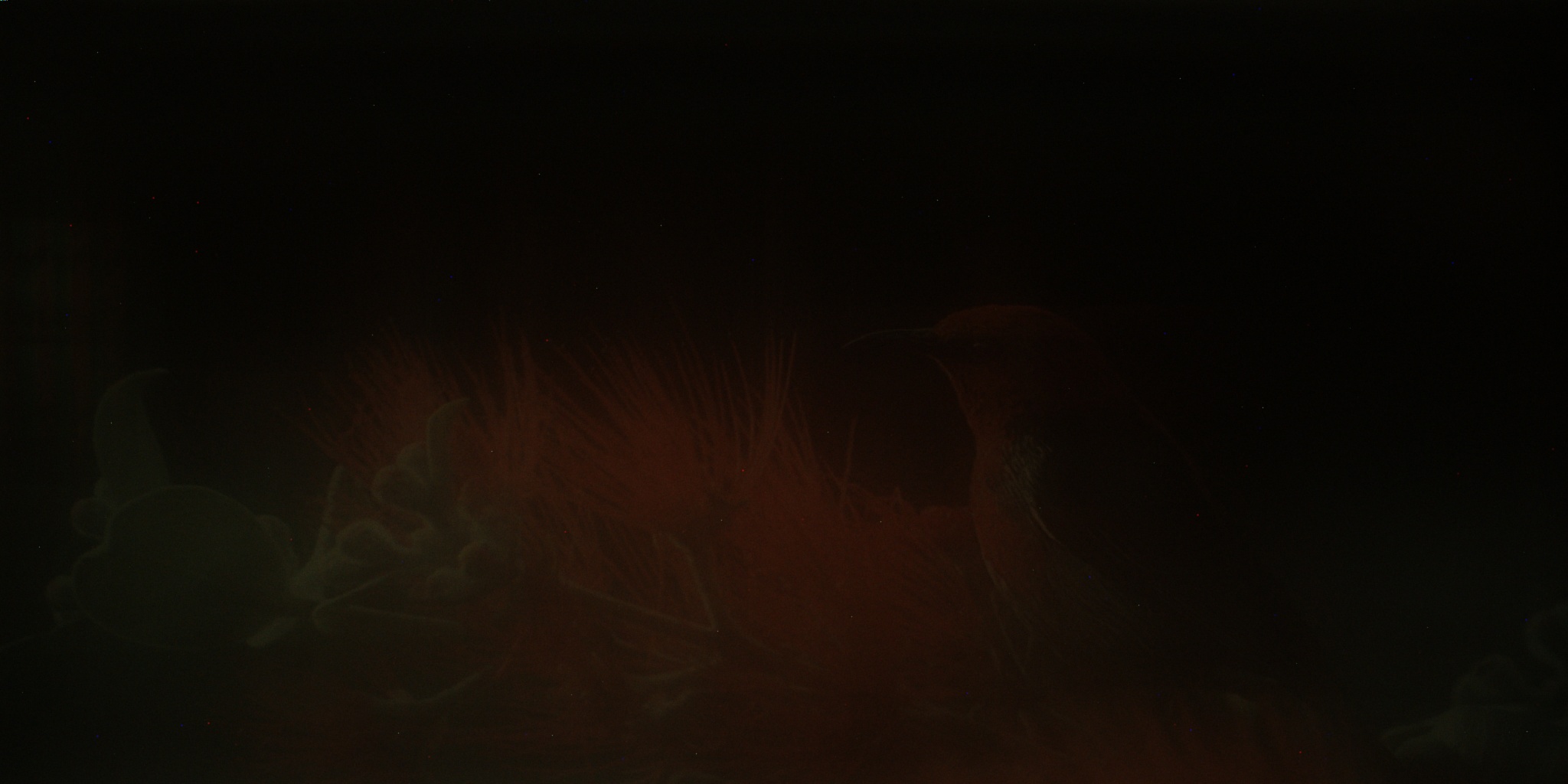}
		\fontsize{8}{12pt}\selectfont PSNR-$20.77dB$/SSIM-$0.75$
		\vskip 2pt
		\includegraphics[scale=\scale]{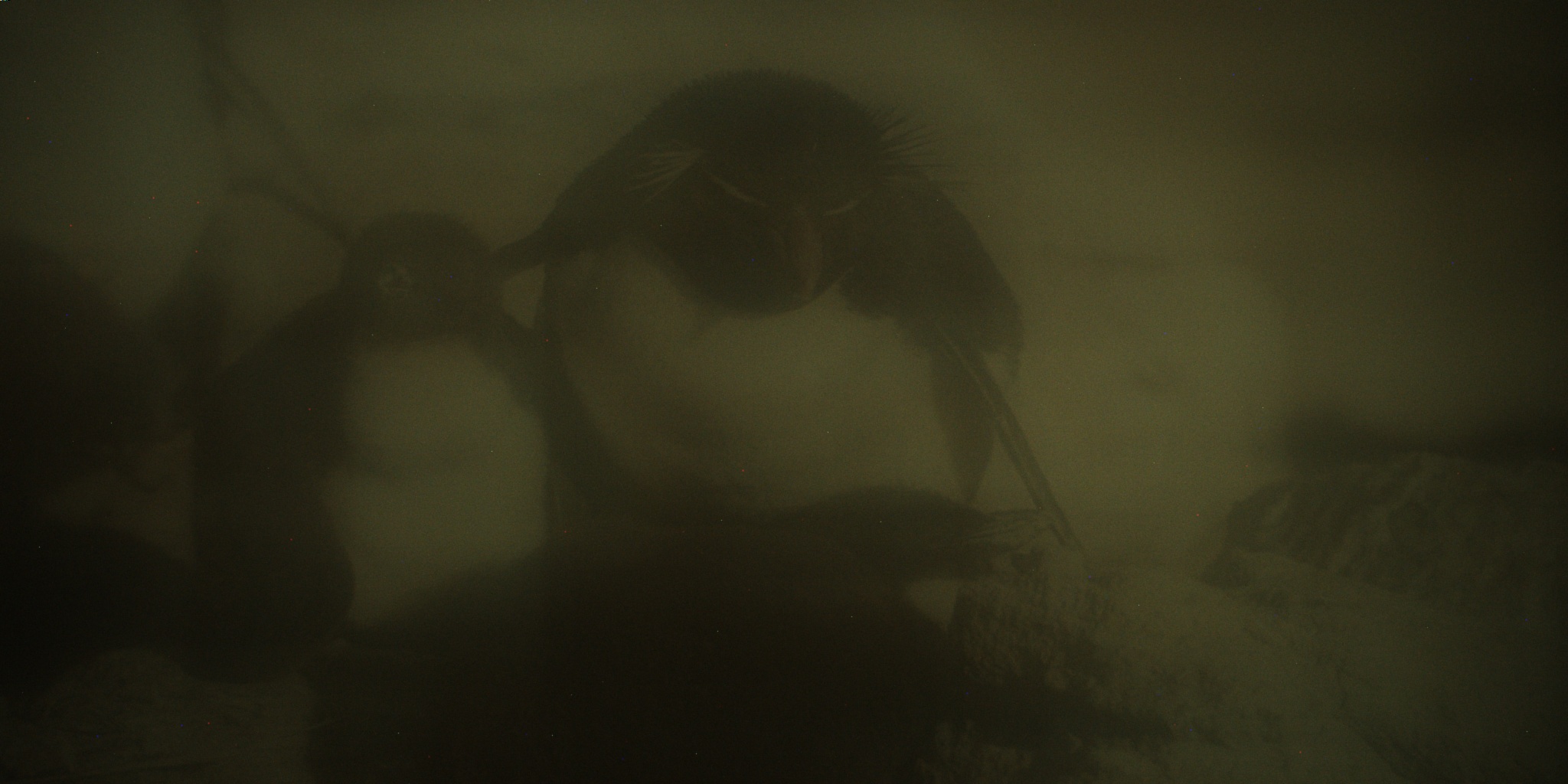}
		\fontsize{8}{12pt}\selectfont PSNR-$11.38dB$/SSIM-$0.58$
		\vskip 2pt
		\includegraphics[scale=\scale]{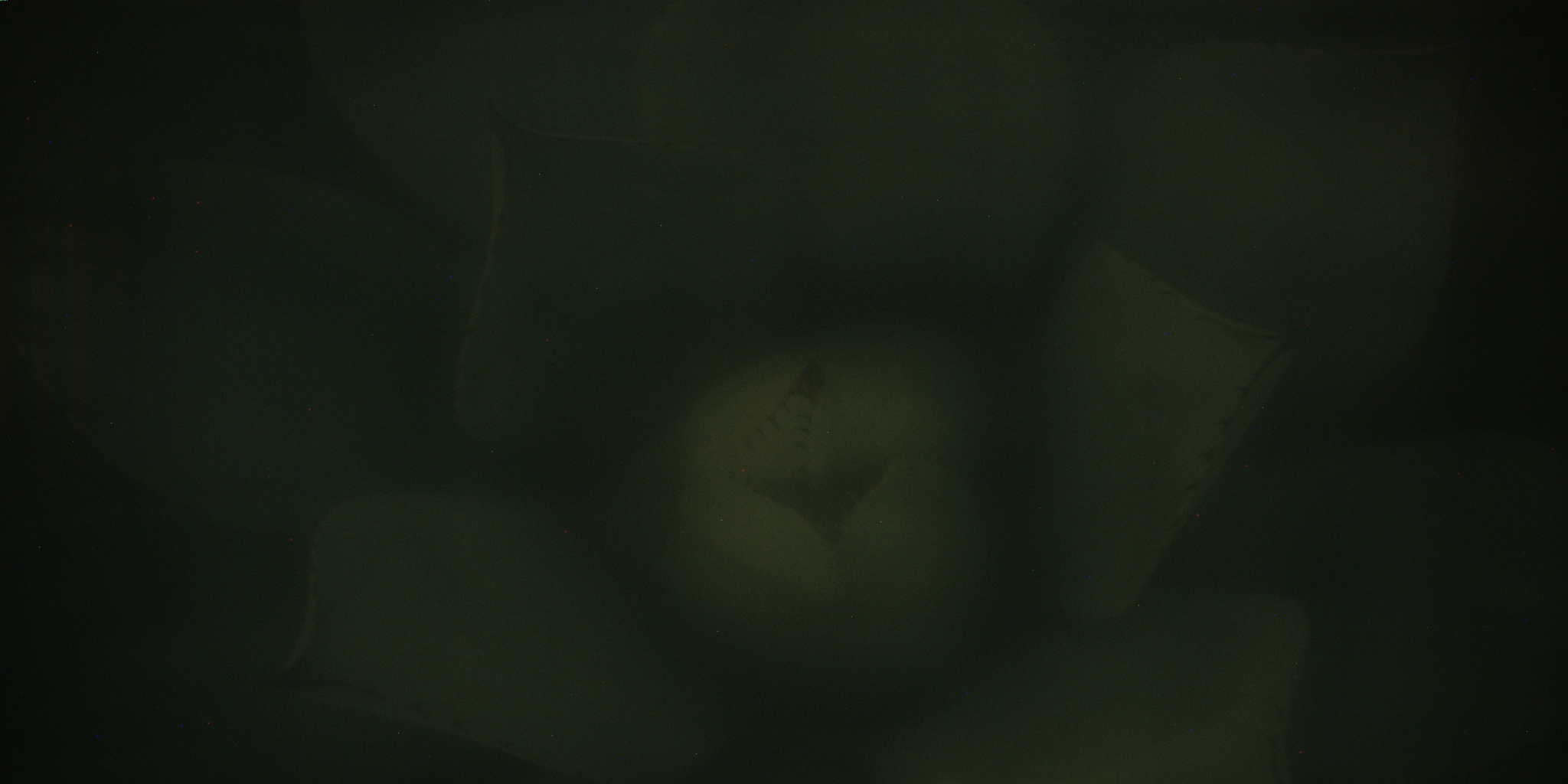}
		\fontsize{8}{12pt}\selectfont PSNR-$15.64dB$/SSIM-$0.66$
		\vskip 2pt
\includegraphics[scale=\scale]{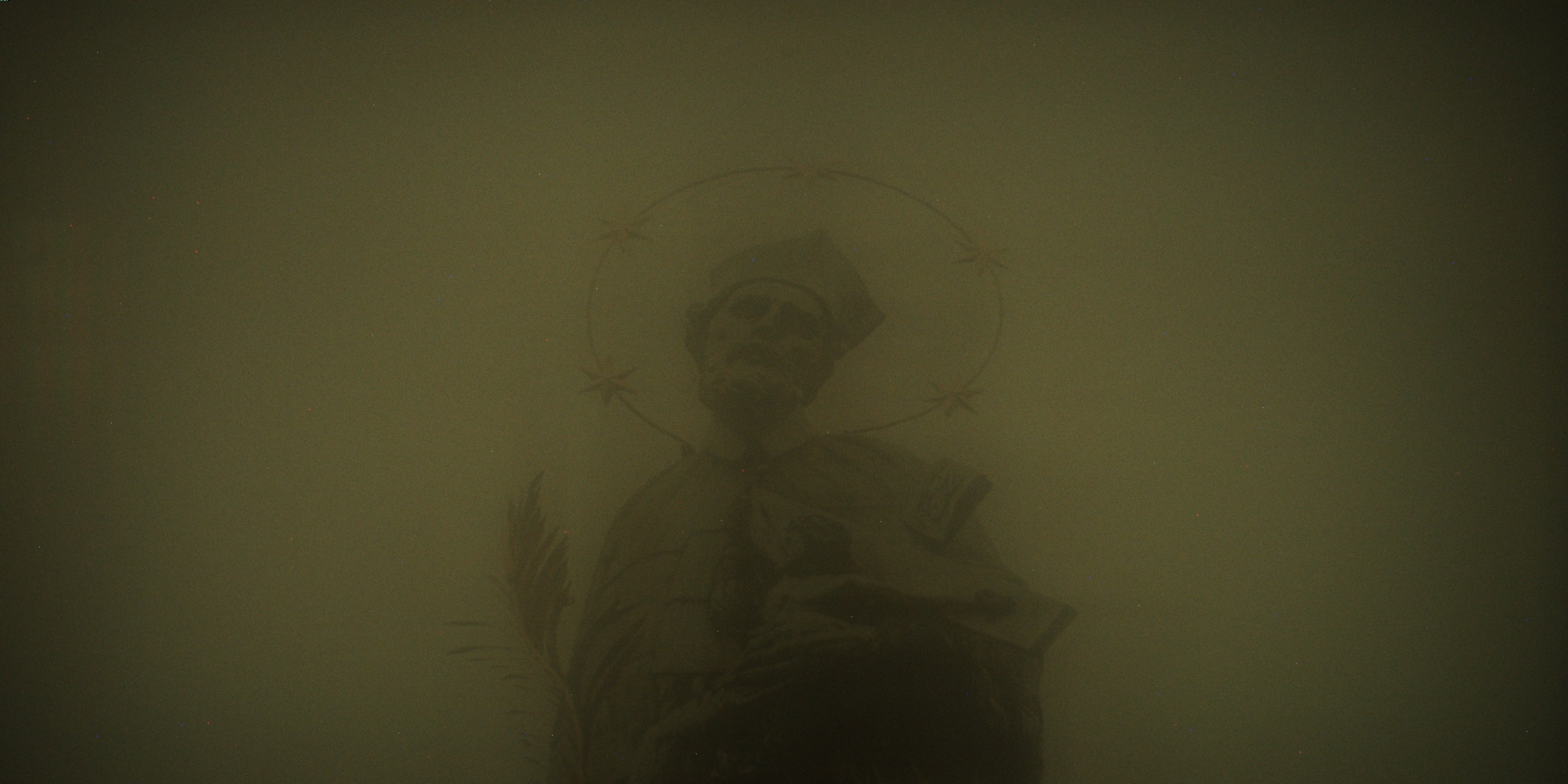}
		\fontsize{8}{12pt}\selectfont PSNR-$9.86dB$/SSIM-$0.58$
		\vskip 2pt
		\includegraphics[scale=\scale]{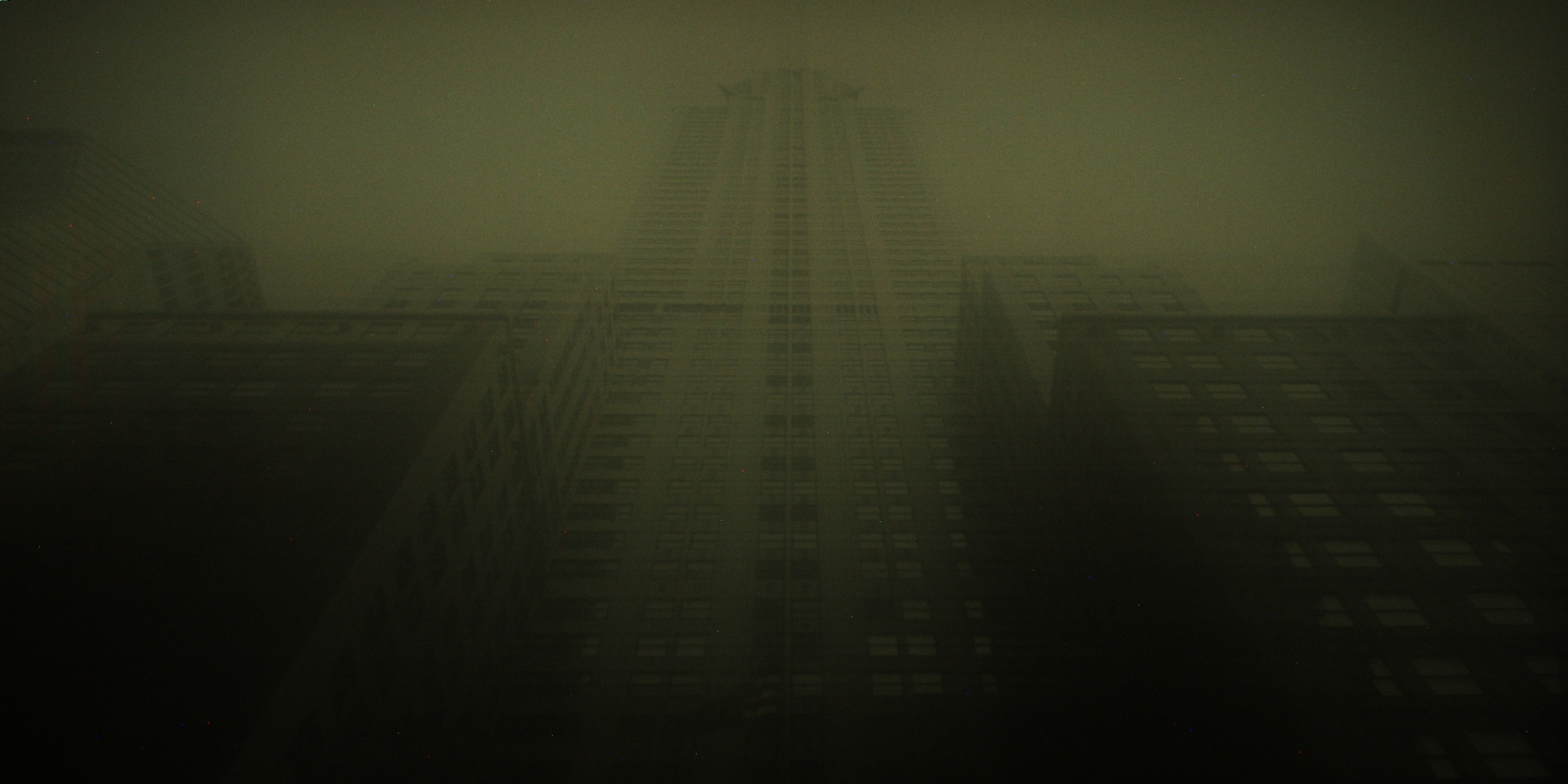}
		\fontsize{8}{12pt}\selectfont PSNR-$10.38dB$/SSIM-$0.54$
		\vskip 2pt
        \fontsize{9}{12pt}\selectfont (a) Input
		\end{center}
  \end{minipage}
  \begin{minipage}{\x\linewidth}
		\begin{center}
		\includegraphics[scale=\scale]{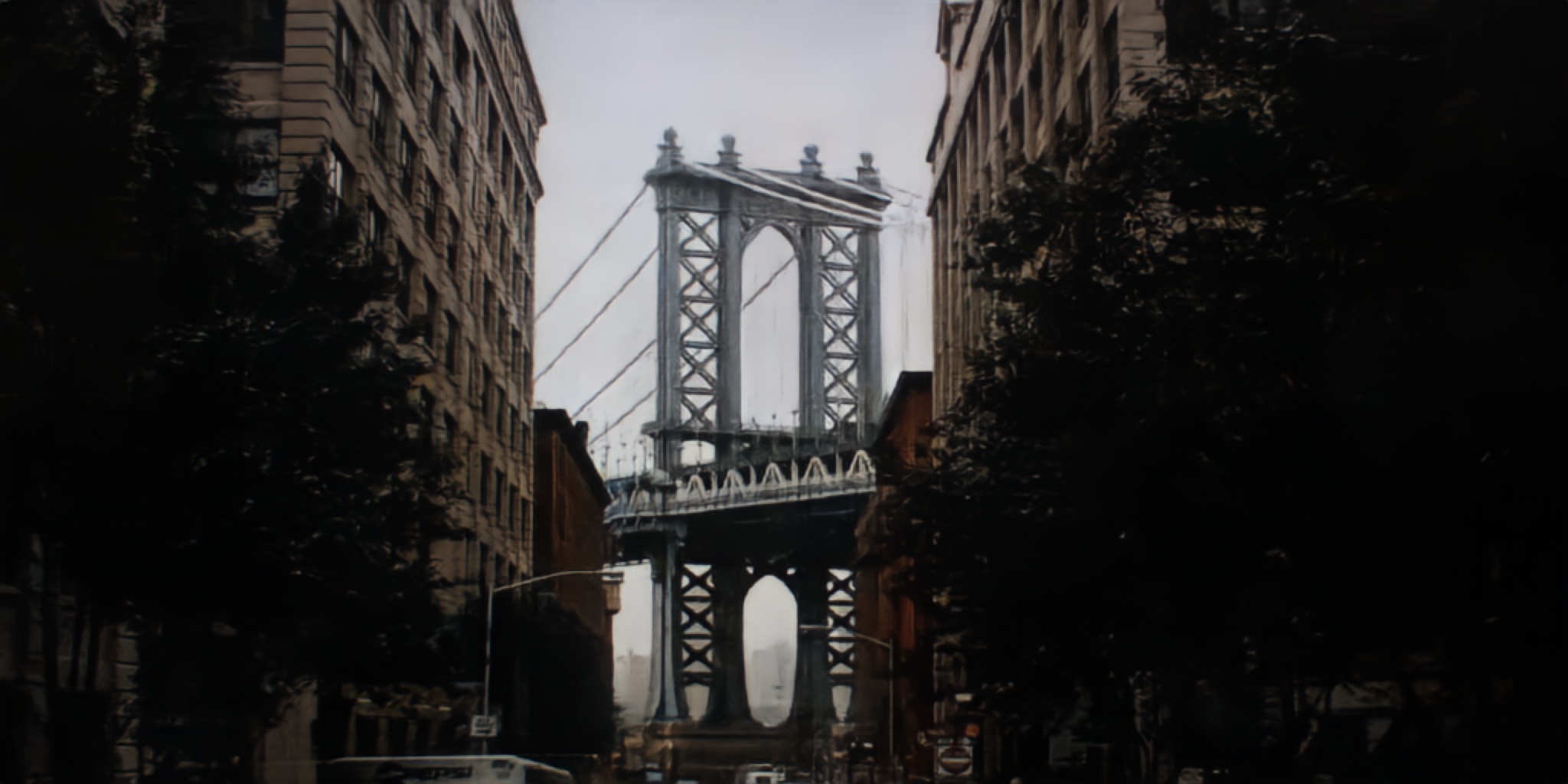}
		\fontsize{8}{12pt}\selectfont PSNR-$32.67 dB$/SSIM-$0.89$
		\vskip 2pt
		\includegraphics[scale=\scale]{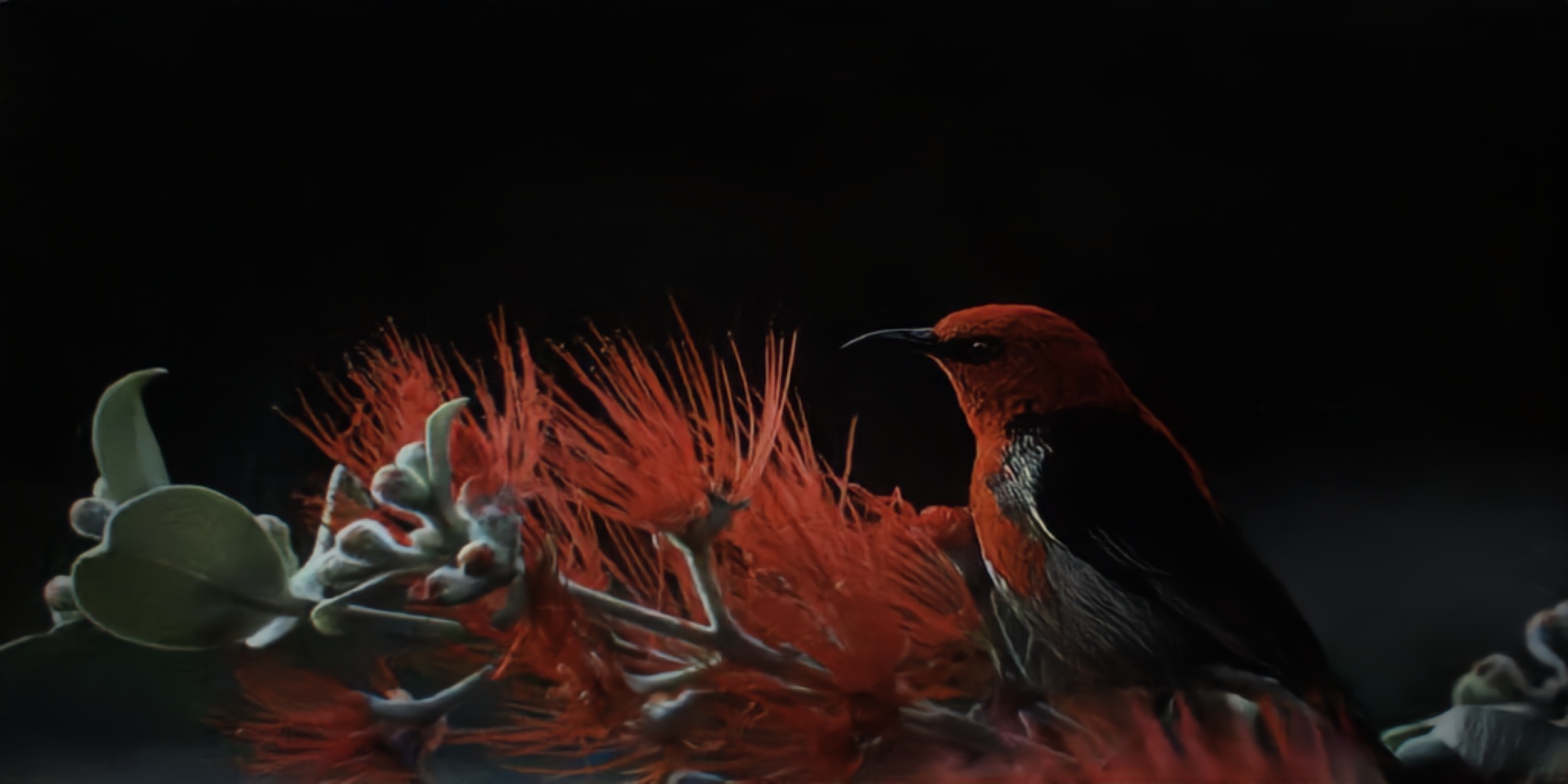}
		\fontsize{8}{12pt}\selectfont PSNR-$35.65 dB$/SSIM-$0.95$
		\vskip 2pt
		\includegraphics[scale=\scale]{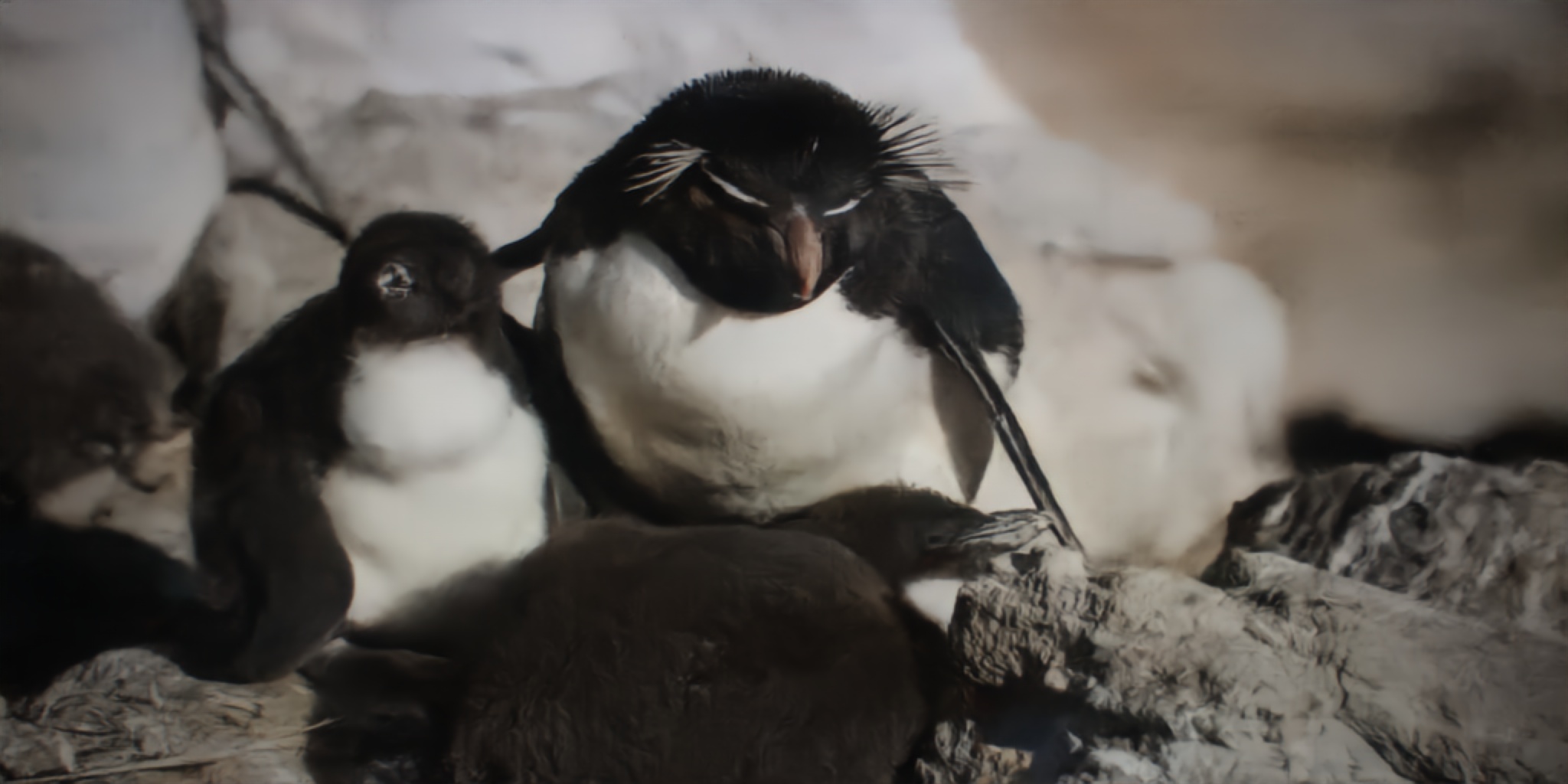}
		\fontsize{8}{12pt}\selectfont PSNR-$33.26 dB$/SSIM-$0.93$
		\vskip 2pt
        \includegraphics[scale=\scale]{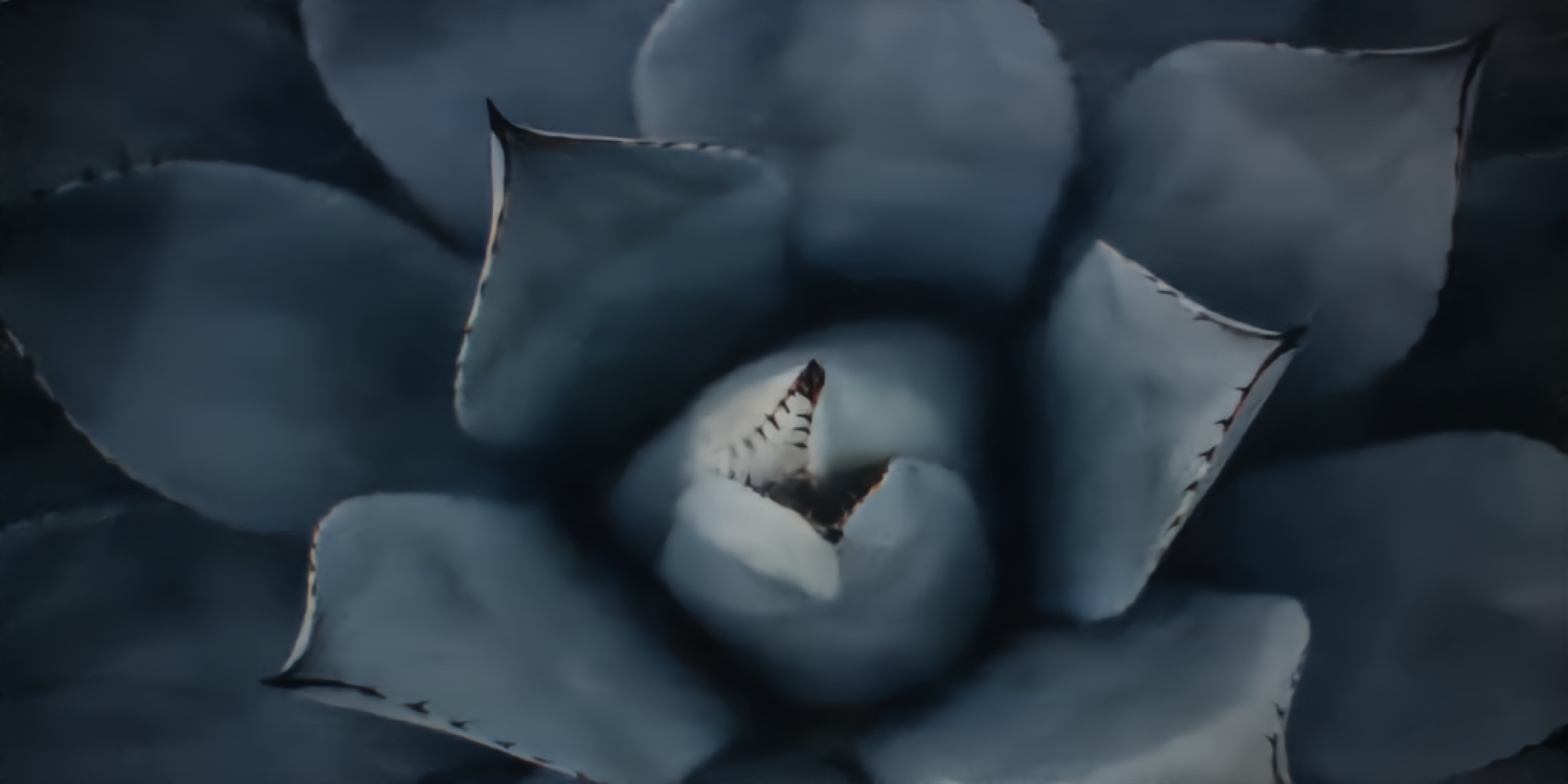}
        \fontsize{8}{12pt}\selectfont PSNR-$37.85 dB$/SSIM-$0.98$
        \vskip 2pt
        \includegraphics[scale=\scale]{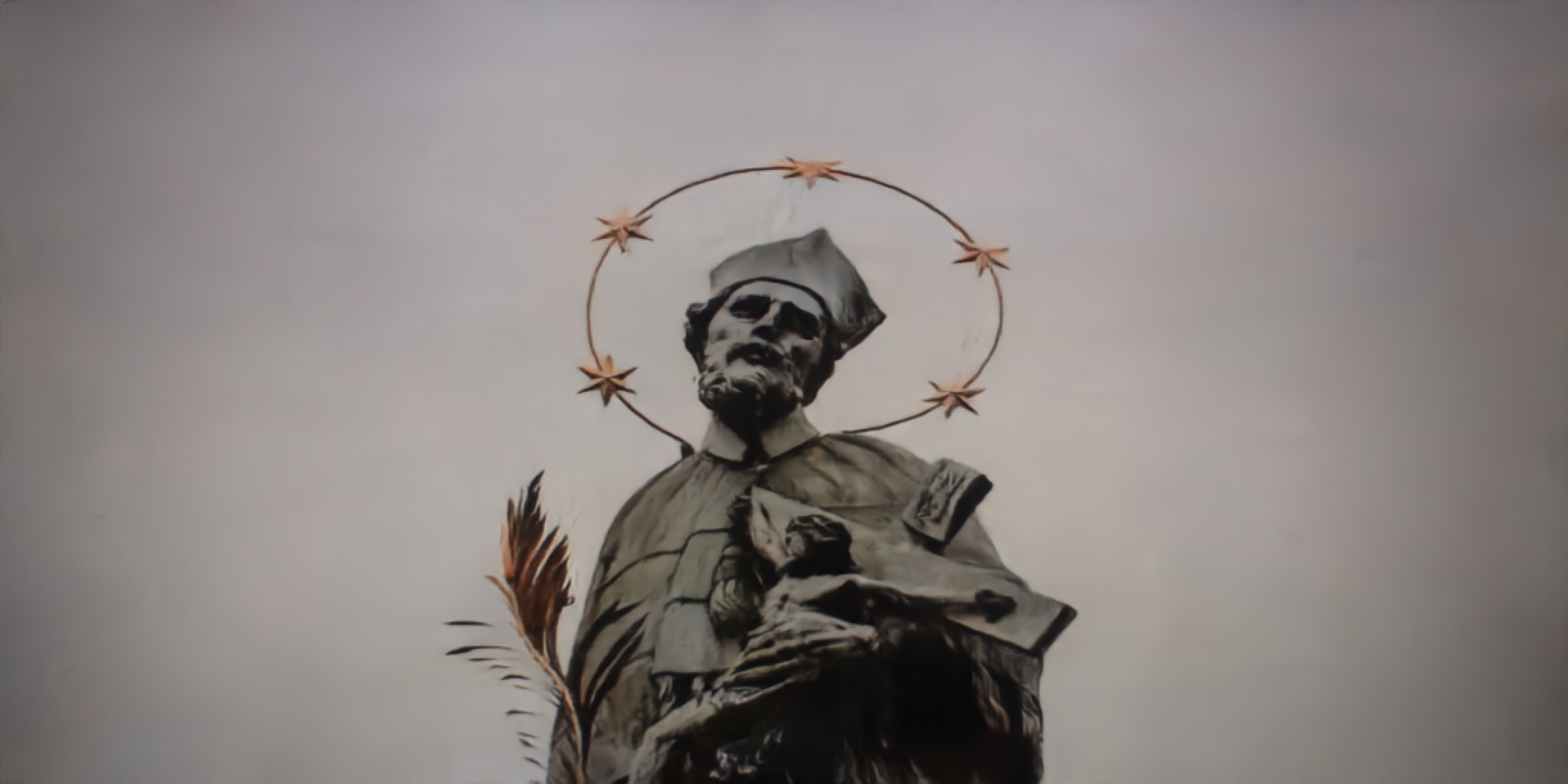}
        \fontsize{8}{12pt}\selectfont PSNR-$35.06 dB$/SSIM-$0.96$
        \vskip 2pt
        \includegraphics[scale=\scale]{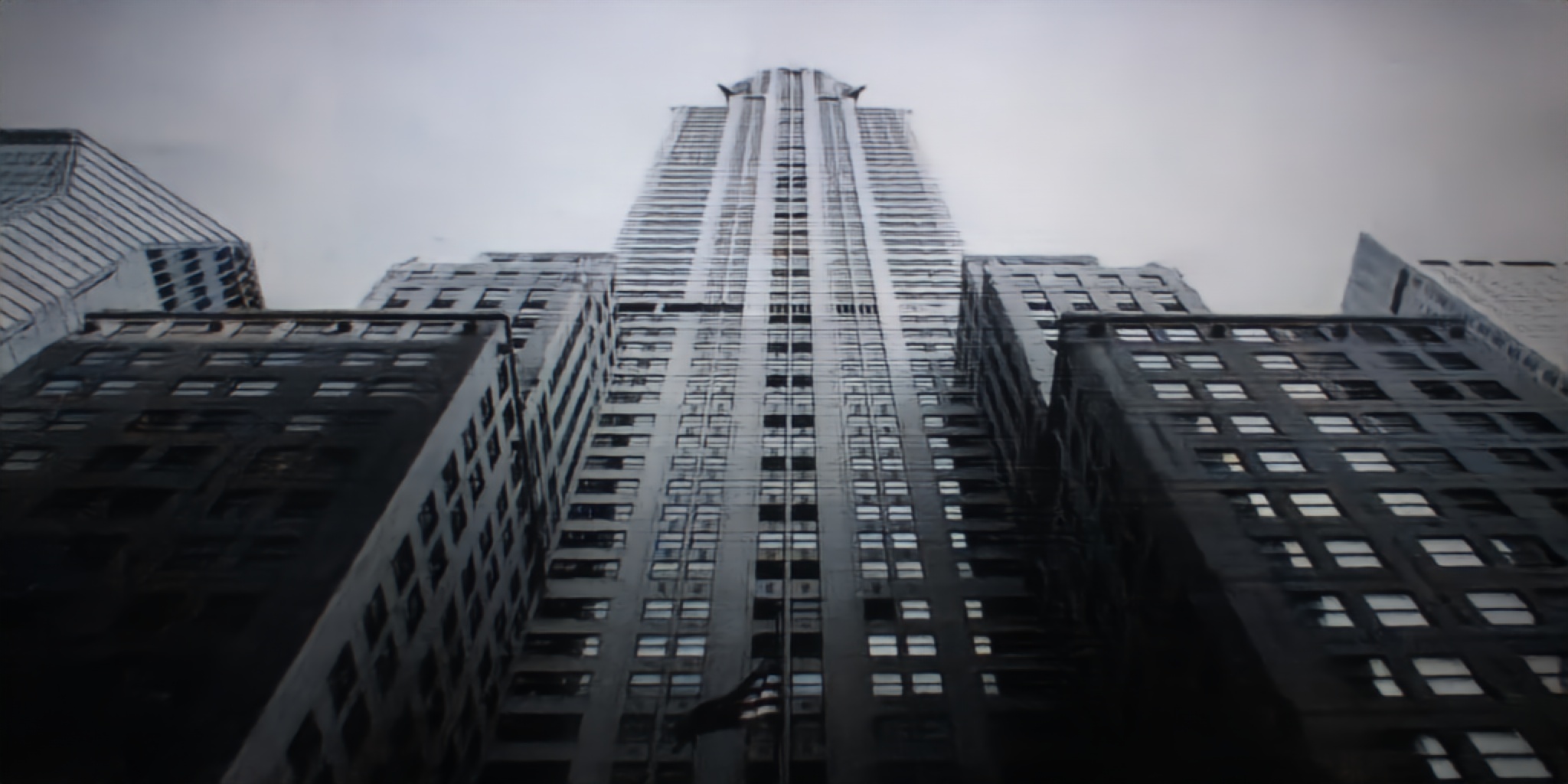}
        \fontsize{8}{12pt}\selectfont PSNR-$31.03 dB$/SSIM-$0.91$
        \vskip 2pt
		\fontsize{9}{12pt}\selectfont (b) Ours
		\end{center}
  \end{minipage}
  \begin{minipage}{\x\linewidth}
		\begin{center}
		\includegraphics[scale=\scale]{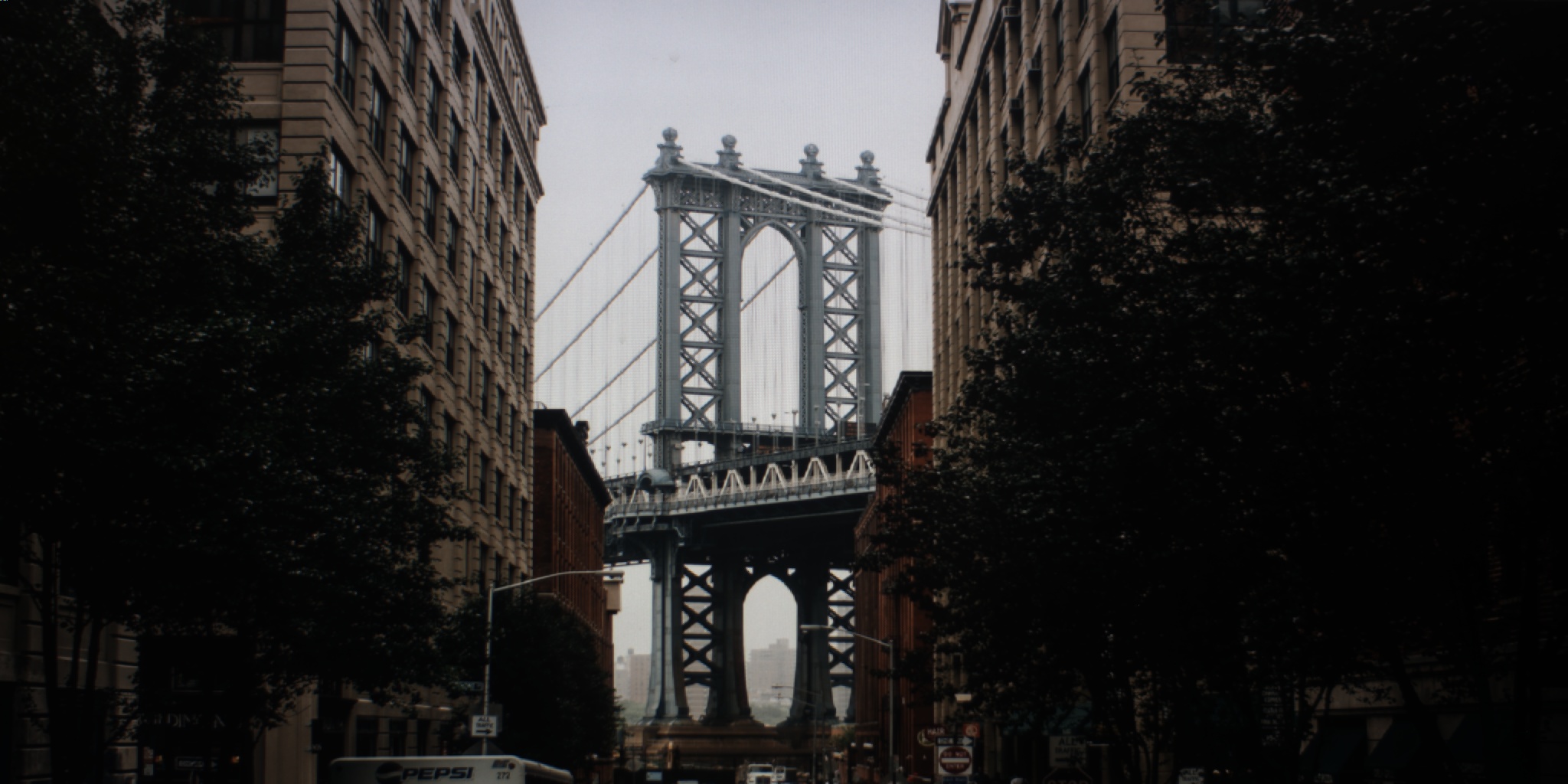}
		\fontsize{8}{12pt}\selectfont {\color{white} PSNR-$\infty  dB$/SSIM-$1$}
		\vskip 2pt
		\includegraphics[scale=\scale]{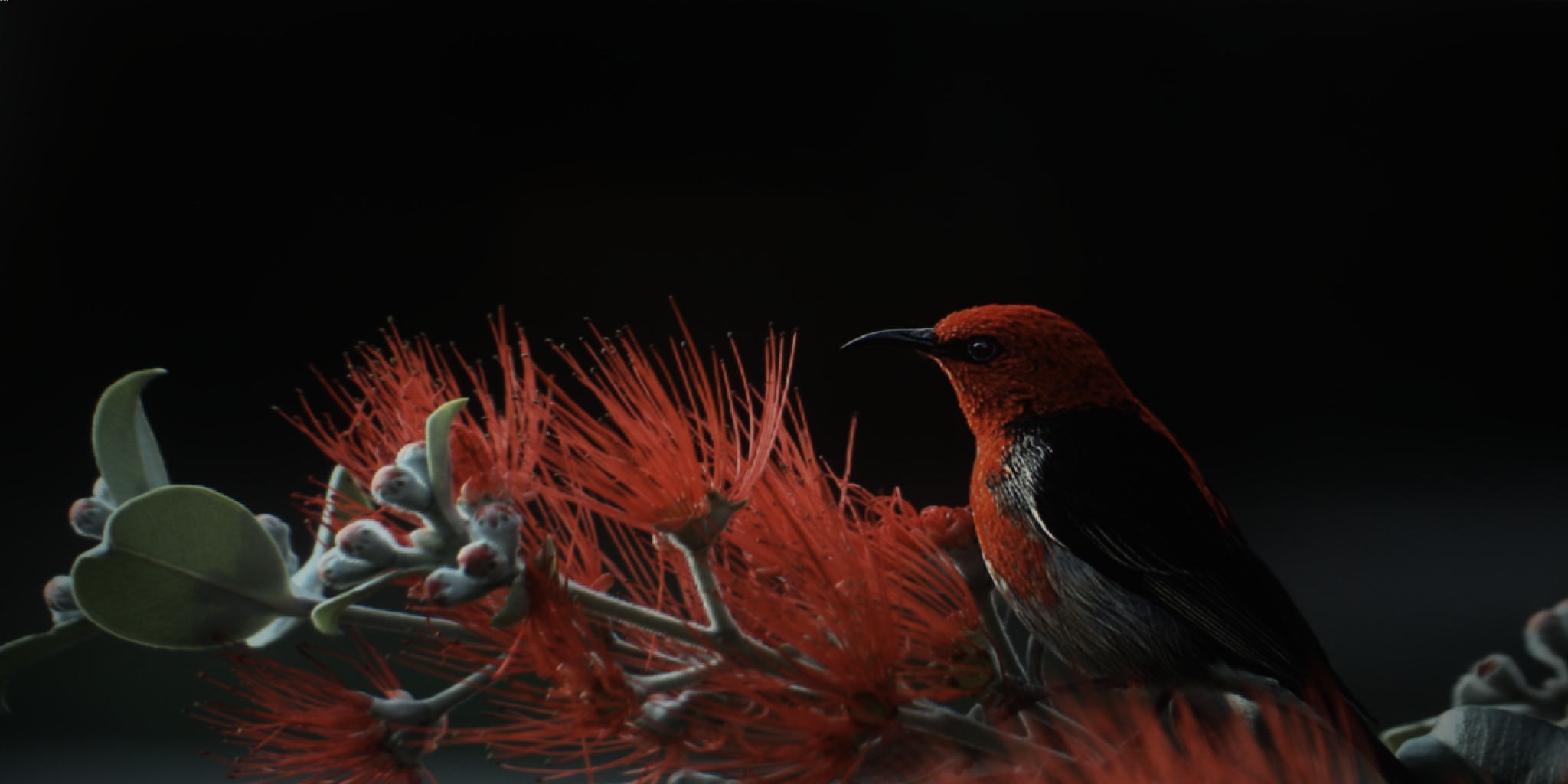}
		\fontsize{8}{12pt}\selectfont {\color{white} PSNR-$\infty  dB$/SSIM-$1$}
		\vskip 2pt
		\includegraphics[scale=\scale]{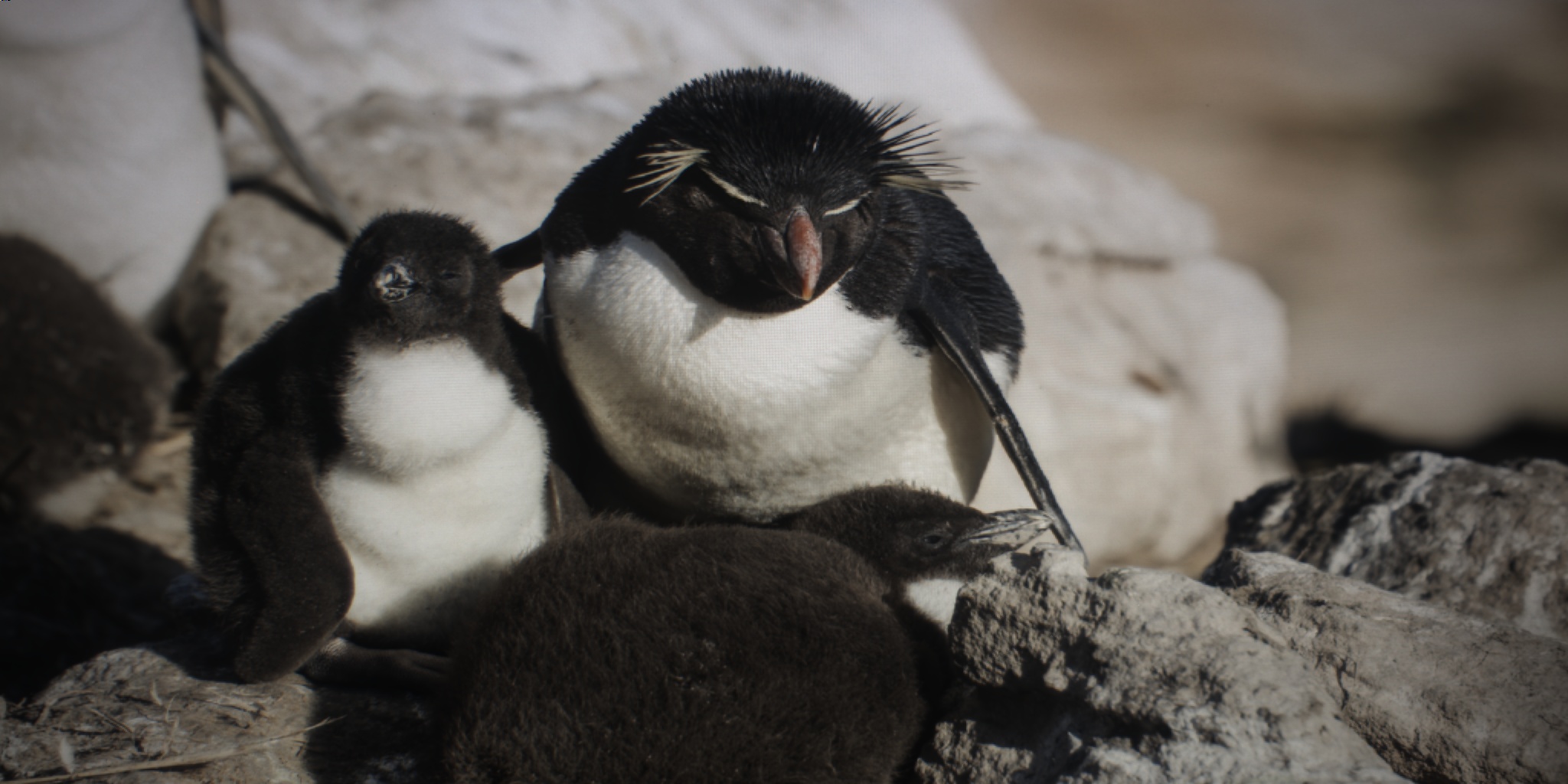}
		\fontsize{8}{12pt}\selectfont {\color{white} PSNR-$\infty  dB$/SSIM-$1$}
		\vskip 2pt
        \includegraphics[scale=\scale]{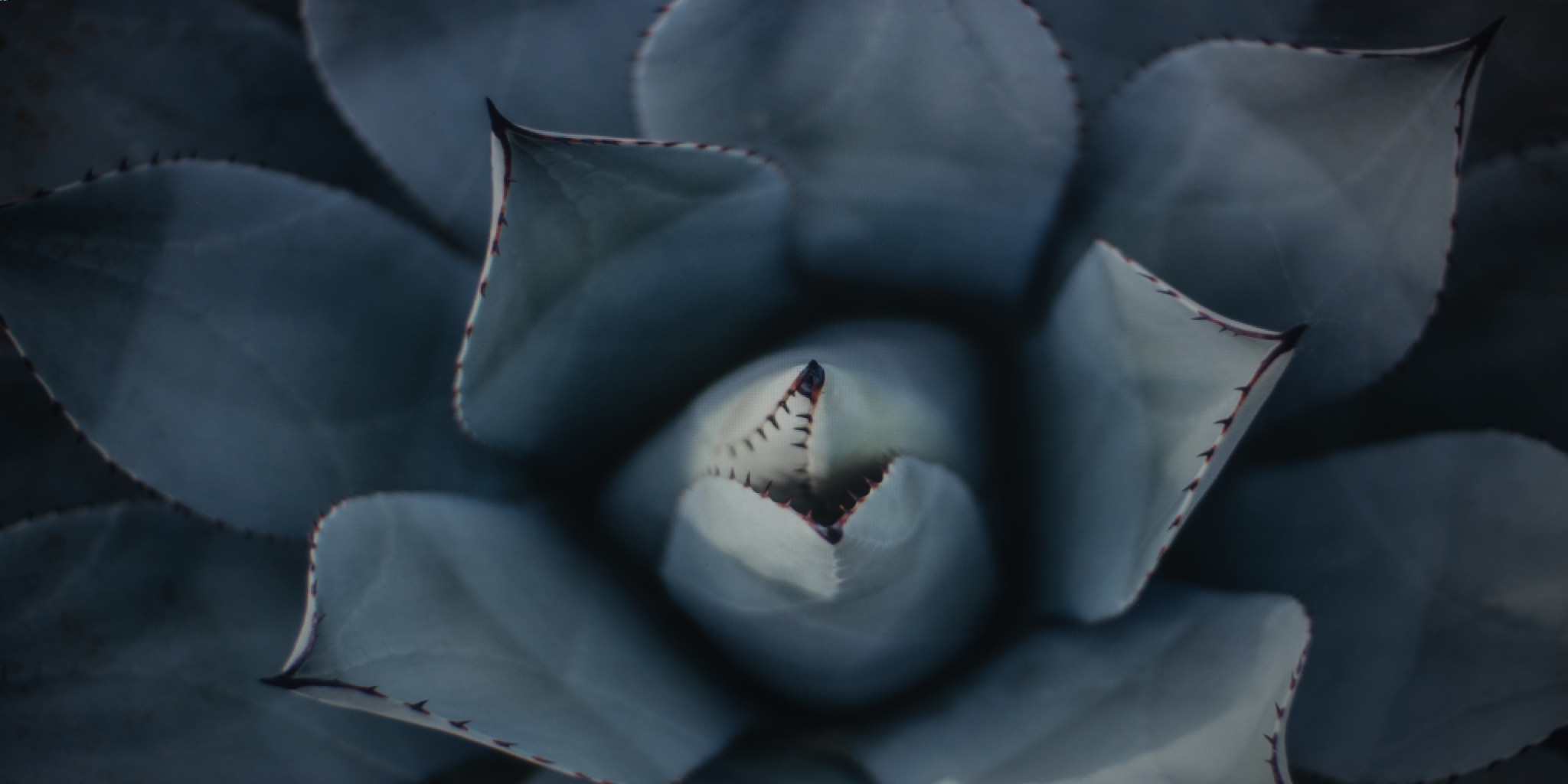}
        \fontsize{8}{12pt}\selectfont {\color{white} PSNR-$\infty  dB$/SSIM-$1$}
        \vskip 2pt
        \includegraphics[scale=\scale]{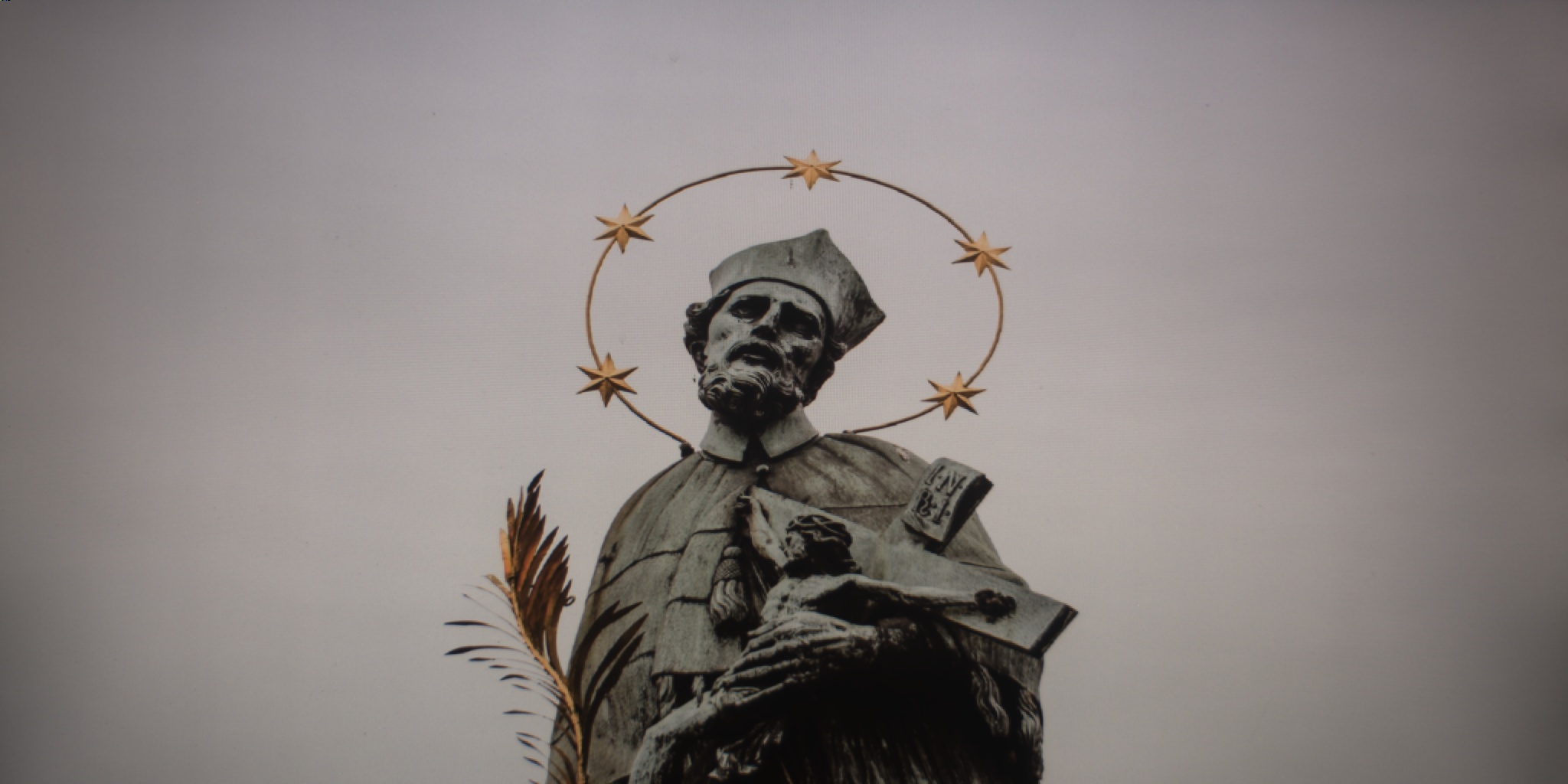}
        \fontsize{8}{12pt}\selectfont {\color{white} PSNR-$\infty  dB$/SSIM-$1$}
        \vskip 2pt
        \includegraphics[scale=\scale]{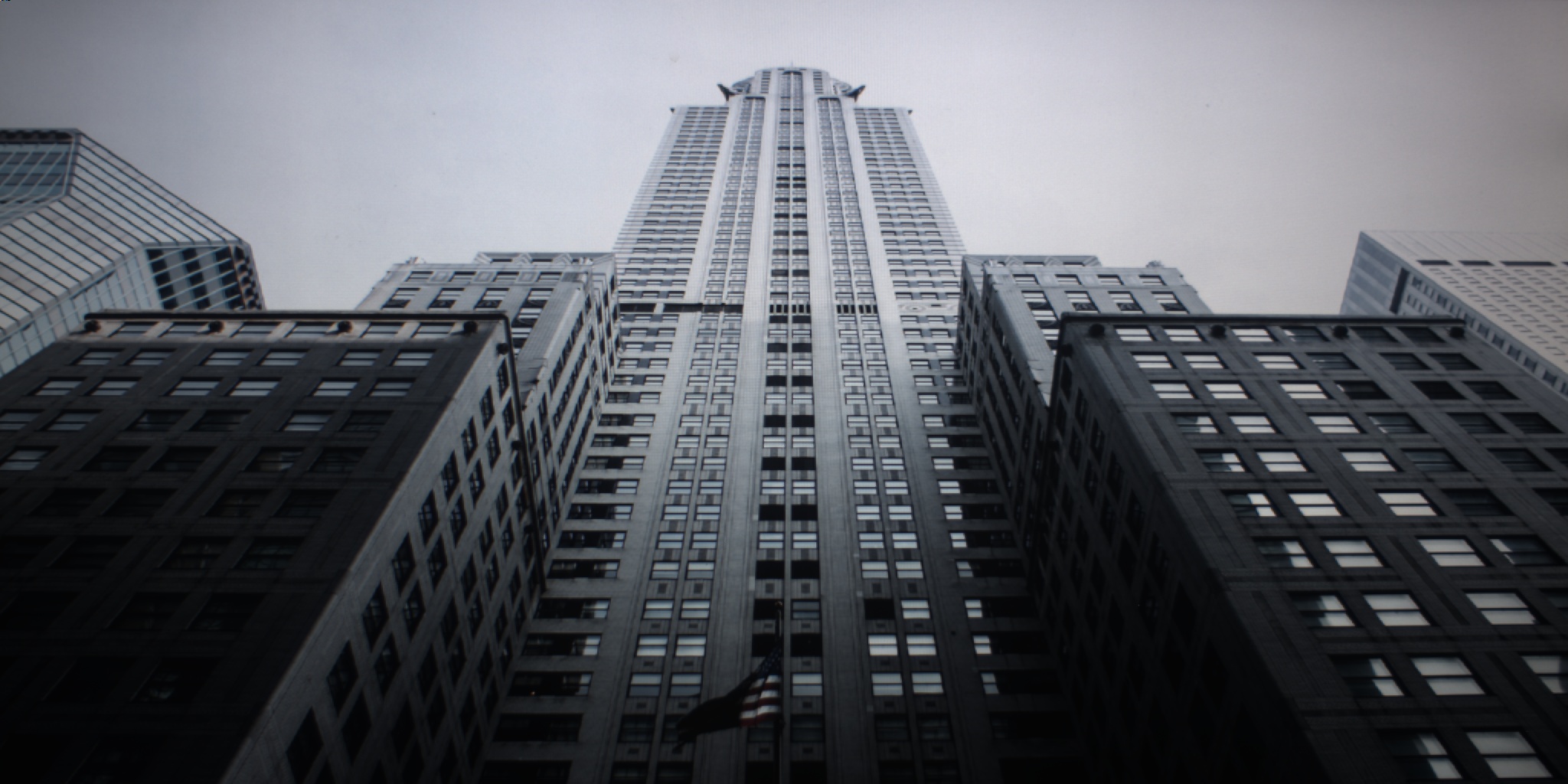}
        \fontsize{8}{12pt}\selectfont {\color{white} PSNR-$\infty  dB$/SSIM-$1$}
        \vskip 2pt
		\fontsize{9}{12pt}\selectfont (c) Ground truth
		\end{center}
  \end{minipage}
\caption{Sample results of the proposed PDCRN for (a). P-OLED degraded images, (b). images restored using PDCRN shows high fidelity with (c). the ground-truth captured without mounting any display.}
\label{fig:poled sample results}
\end{figure}
\begin{figure}
\centering
\newcommand\x{0.3}
\newcommand\scale{0.050}
  \begin{minipage}{\x\linewidth}
		\begin{center}
		\includegraphics[scale=\scale]{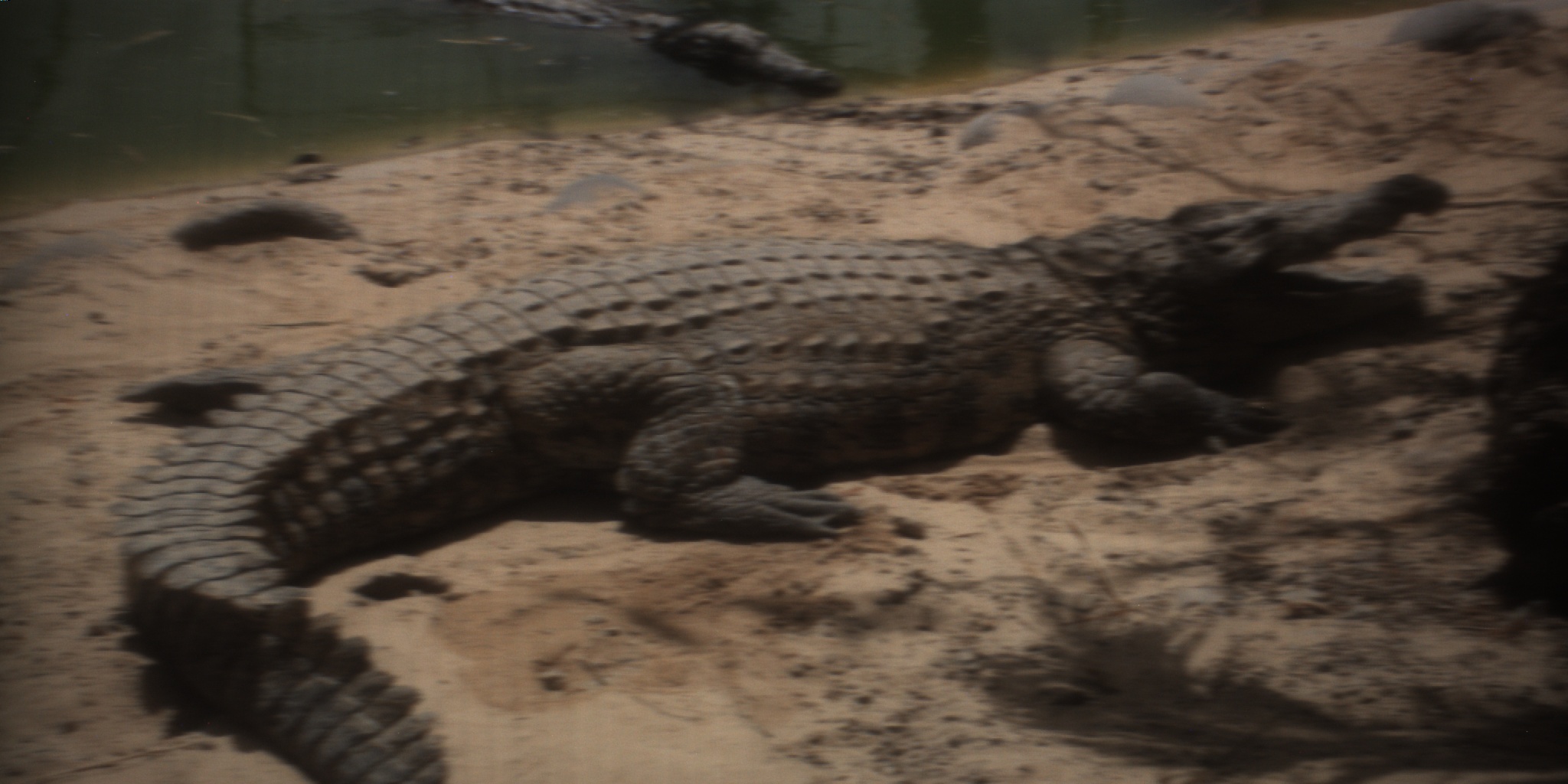}
		\fontsize{8}{12pt}\selectfont PSNR-$27.55dB$/SSIM-$0.77$
		\vskip 2pt
        \includegraphics[scale=\scale]{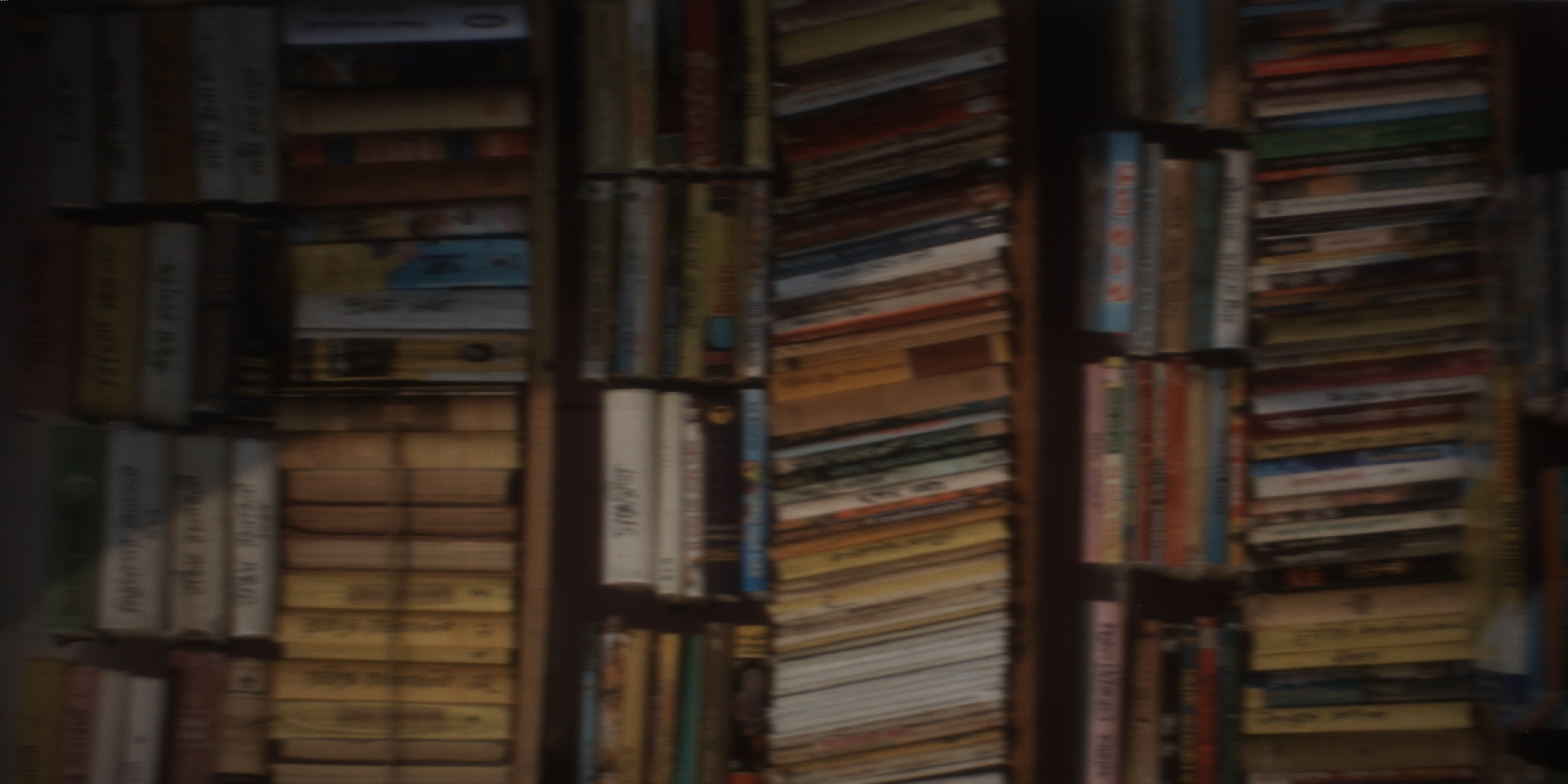}
        \fontsize{8}{12pt}\selectfont PSNR-$29.43dB$/SSIM-$0.84$
        \vskip 2pt
        \includegraphics[scale=\scale]{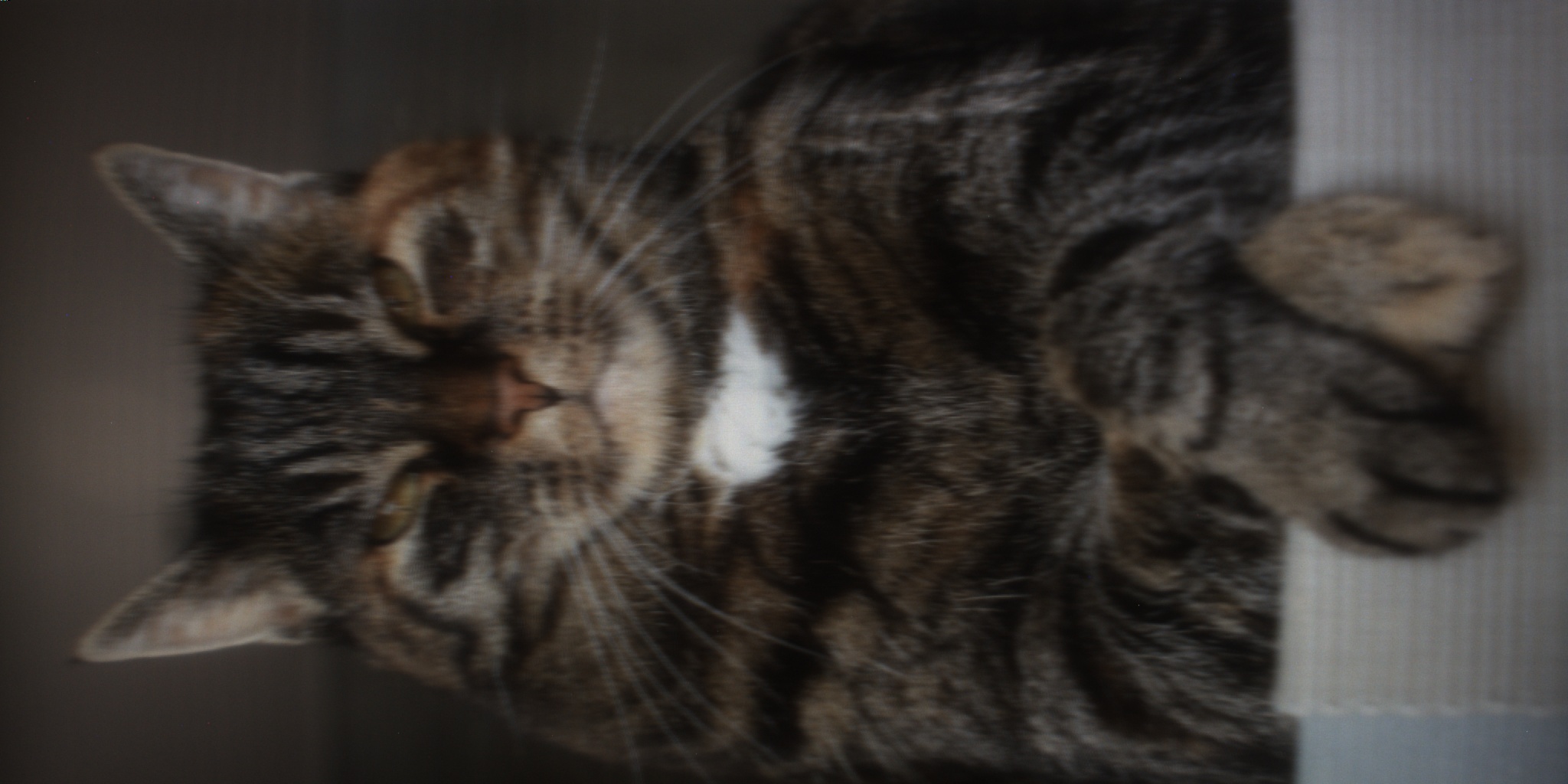}
		\fontsize{8}{12pt}\selectfont PSNR-$28.05dB$/SSIM-$0.77$
		\vskip 2pt
        \includegraphics[scale=\scale]{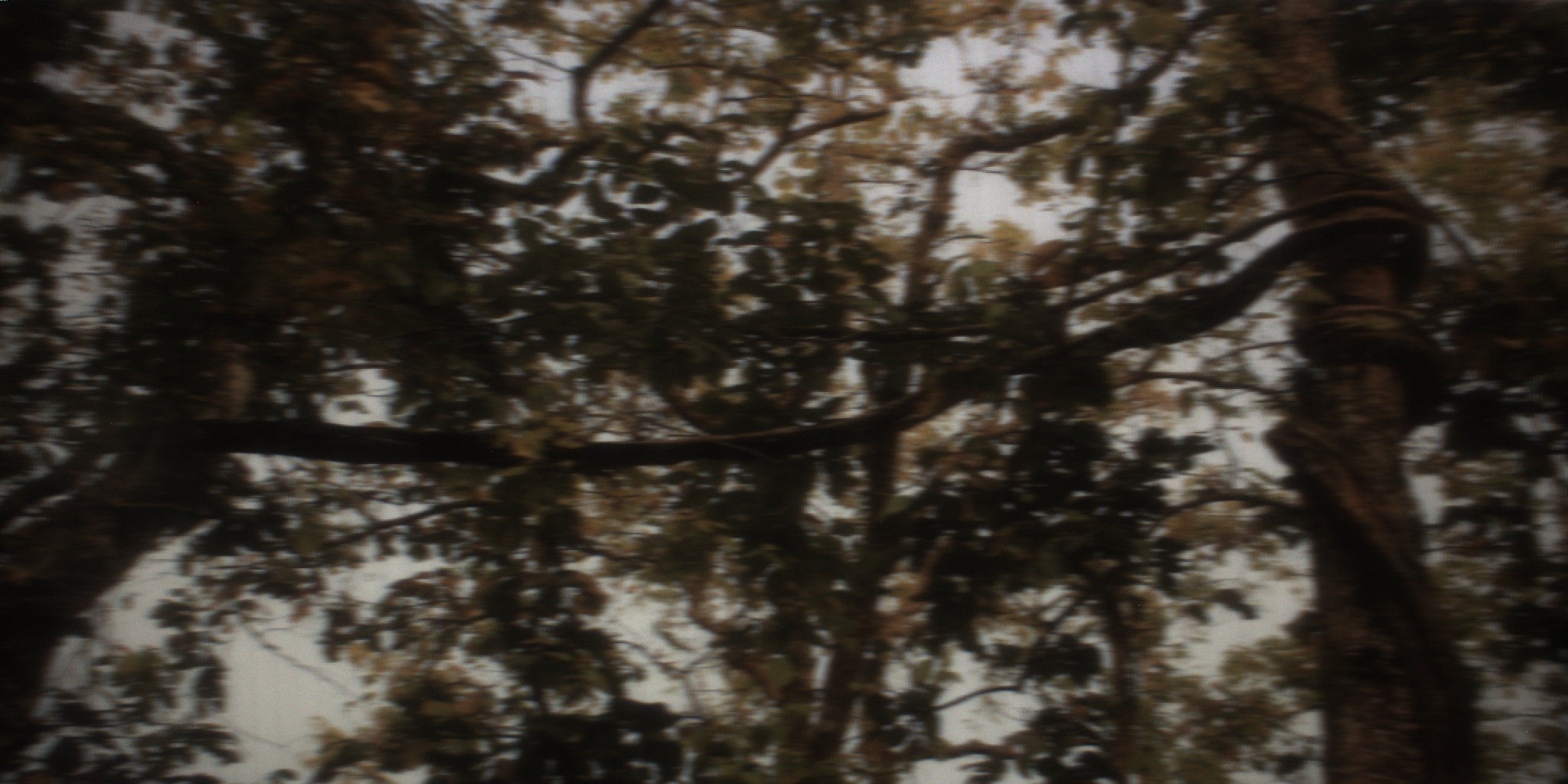}
        \fontsize{8}{12pt}\selectfont PSNR-$21.87dB$/SSIM-$0.68$
        \vskip 2pt
        \includegraphics[scale=\scale]{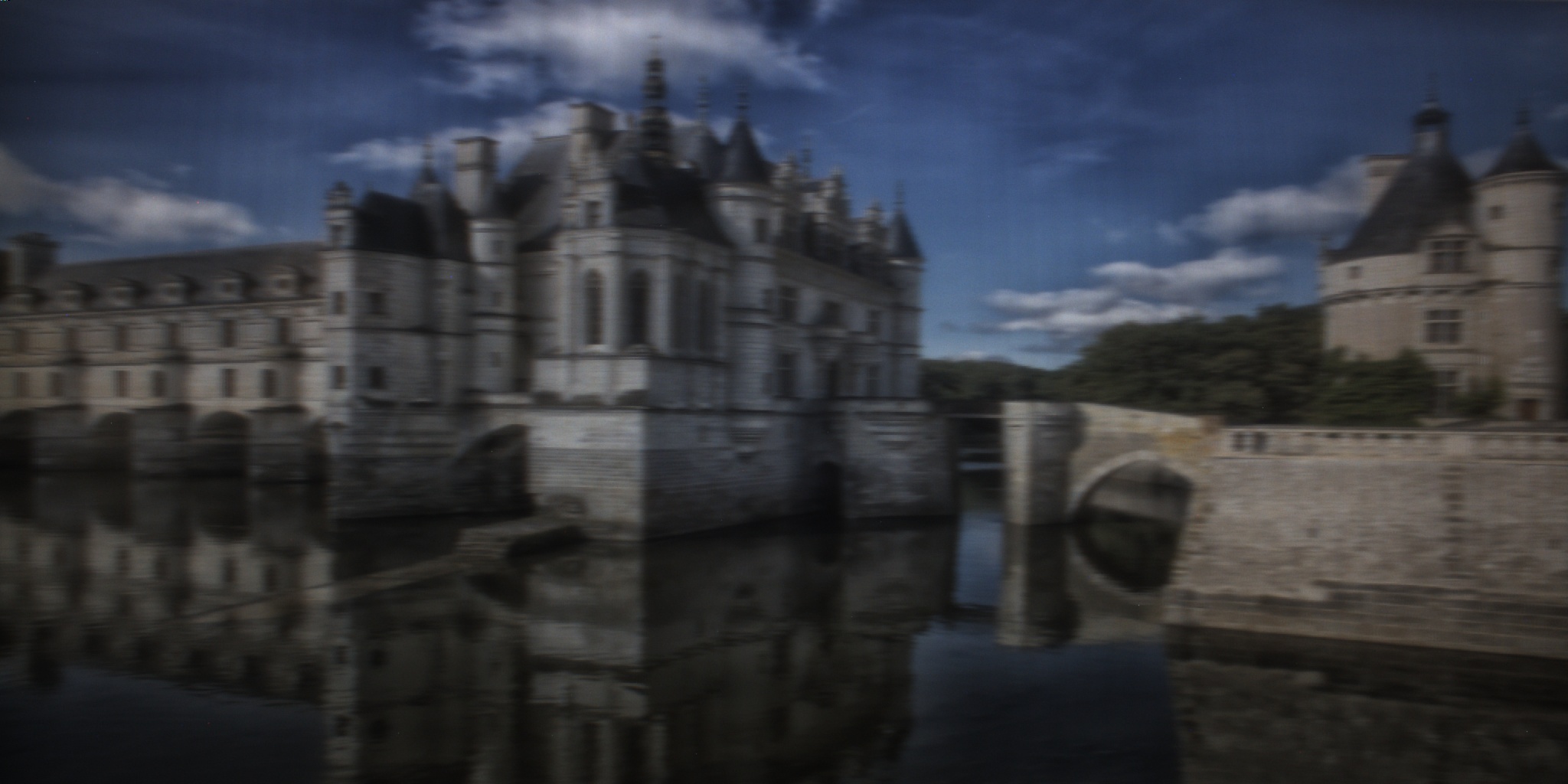}
        \fontsize{8}{12pt}\selectfont PSNR-$29.69dB$/SSIM-$0.87$
        \vskip 2pt
        \includegraphics[scale=\scale]{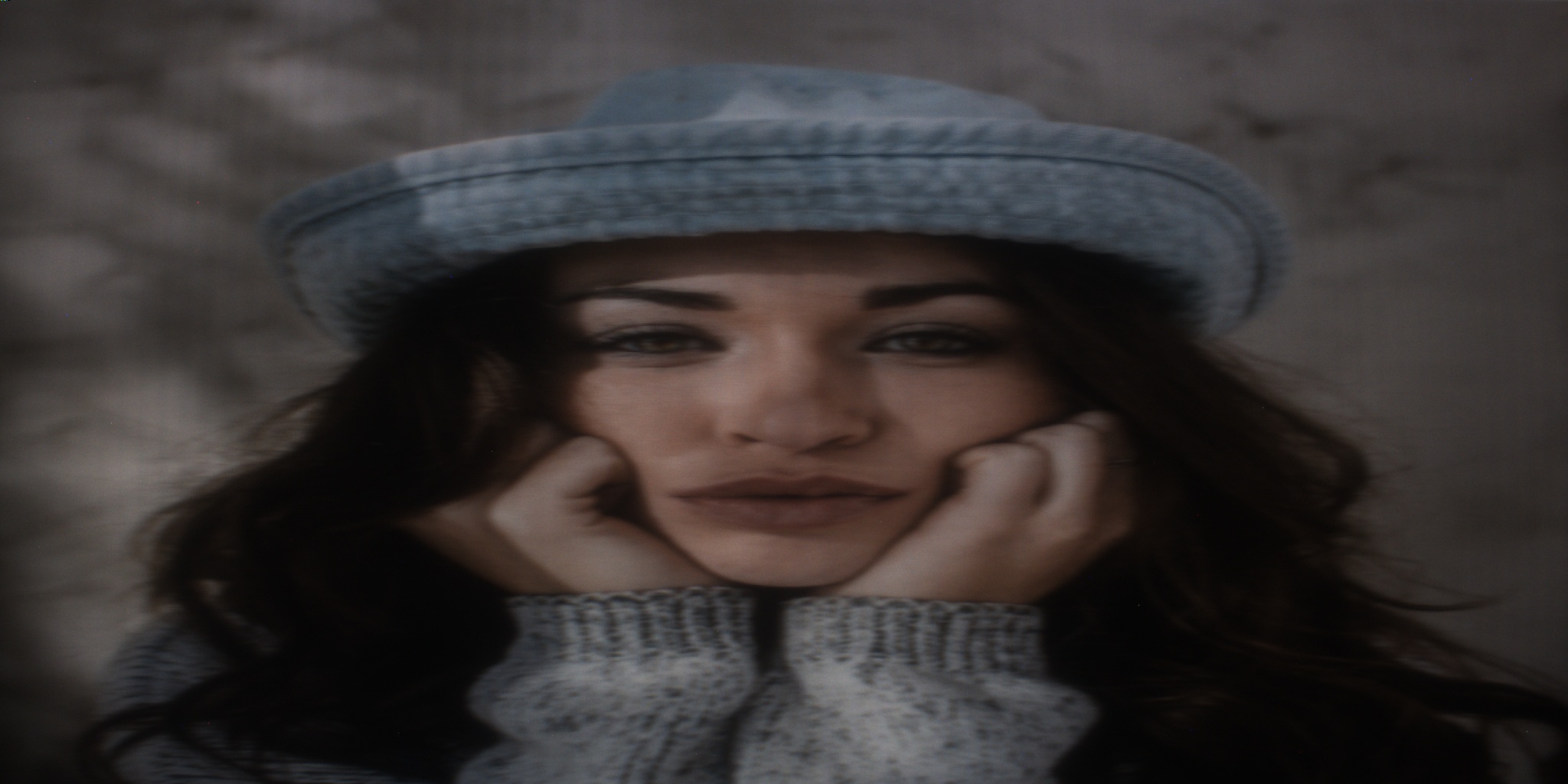}
        \fontsize{8}{12pt}\selectfont PSNR-$31.27dB$/SSIM-$0.88$
        \vskip 2pt
        \fontsize{9}{12pt}\selectfont (a) Input
		\end{center}
  \end{minipage}
  \begin{minipage}{\x\linewidth}
		\begin{center}
		\includegraphics[scale=\scale]{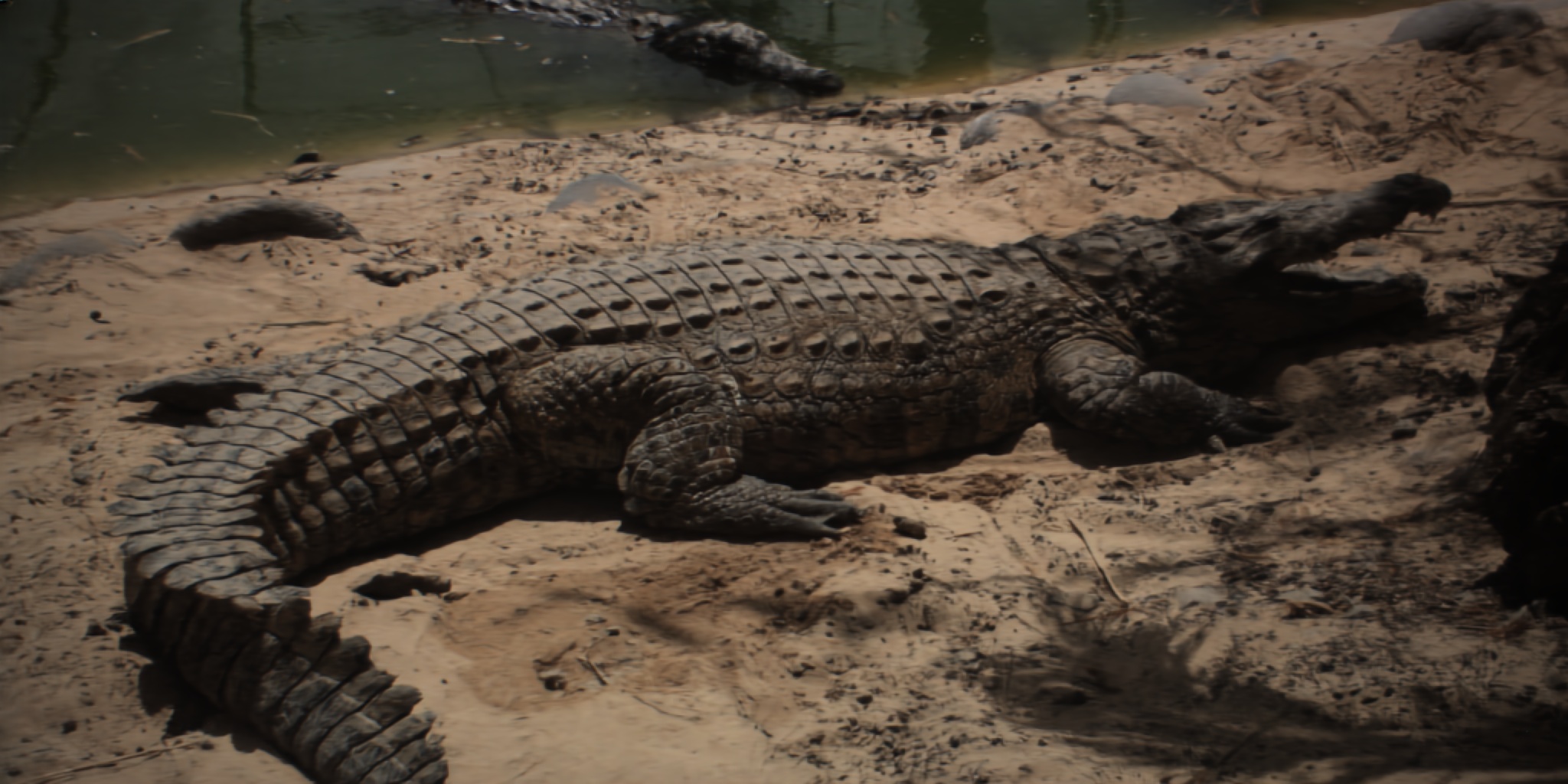}
		\fontsize{8}{12pt}\selectfont PSNR-$34.33 dB$/SSIM-$0.91$
		\vskip 2pt
        \includegraphics[scale=\scale]{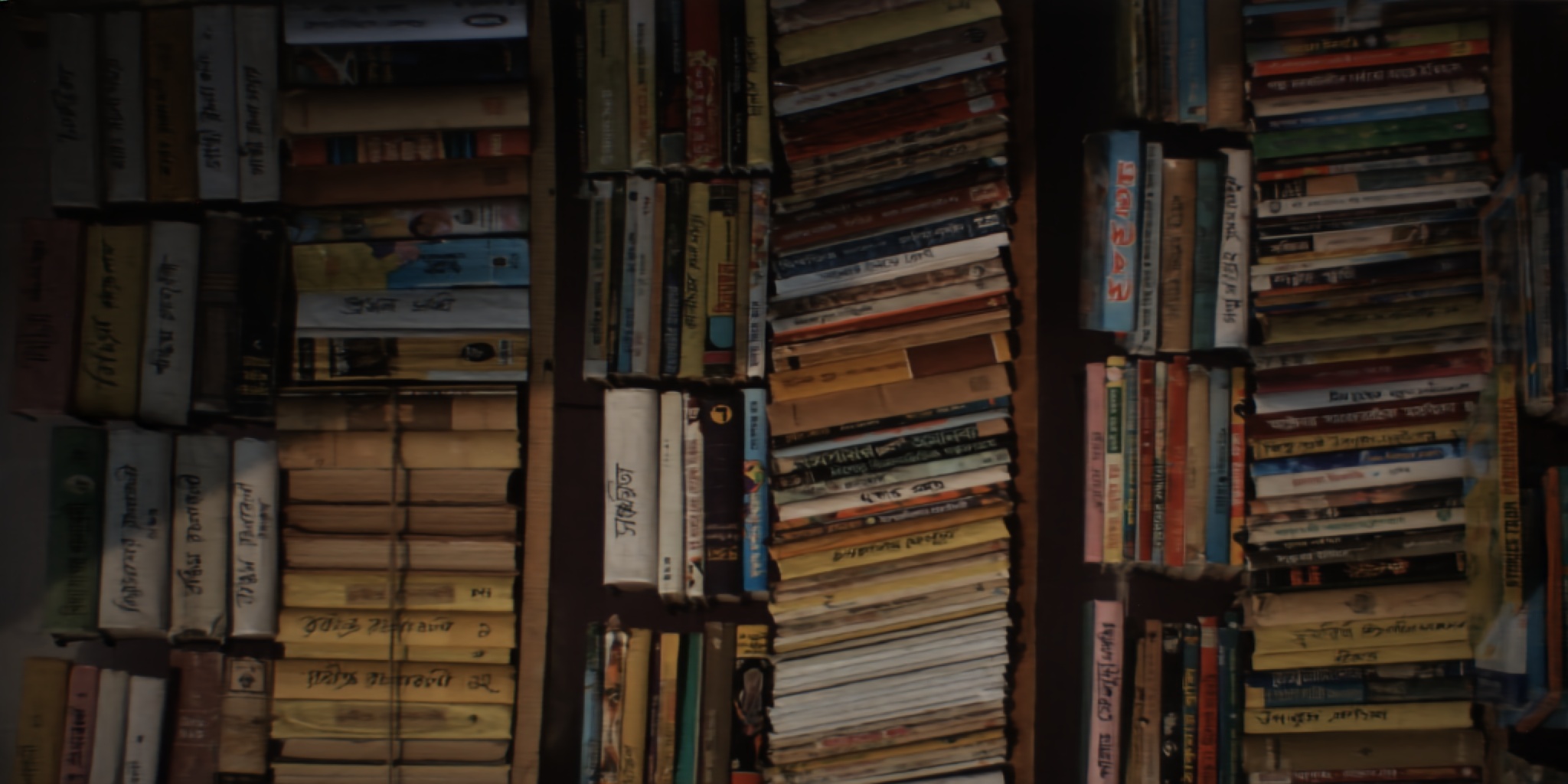}
        \fontsize{8}{12pt}\selectfont PSNR-$38.17 dB$/SSIM-$0.96$
        \vskip 2pt
        \includegraphics[scale=\scale]{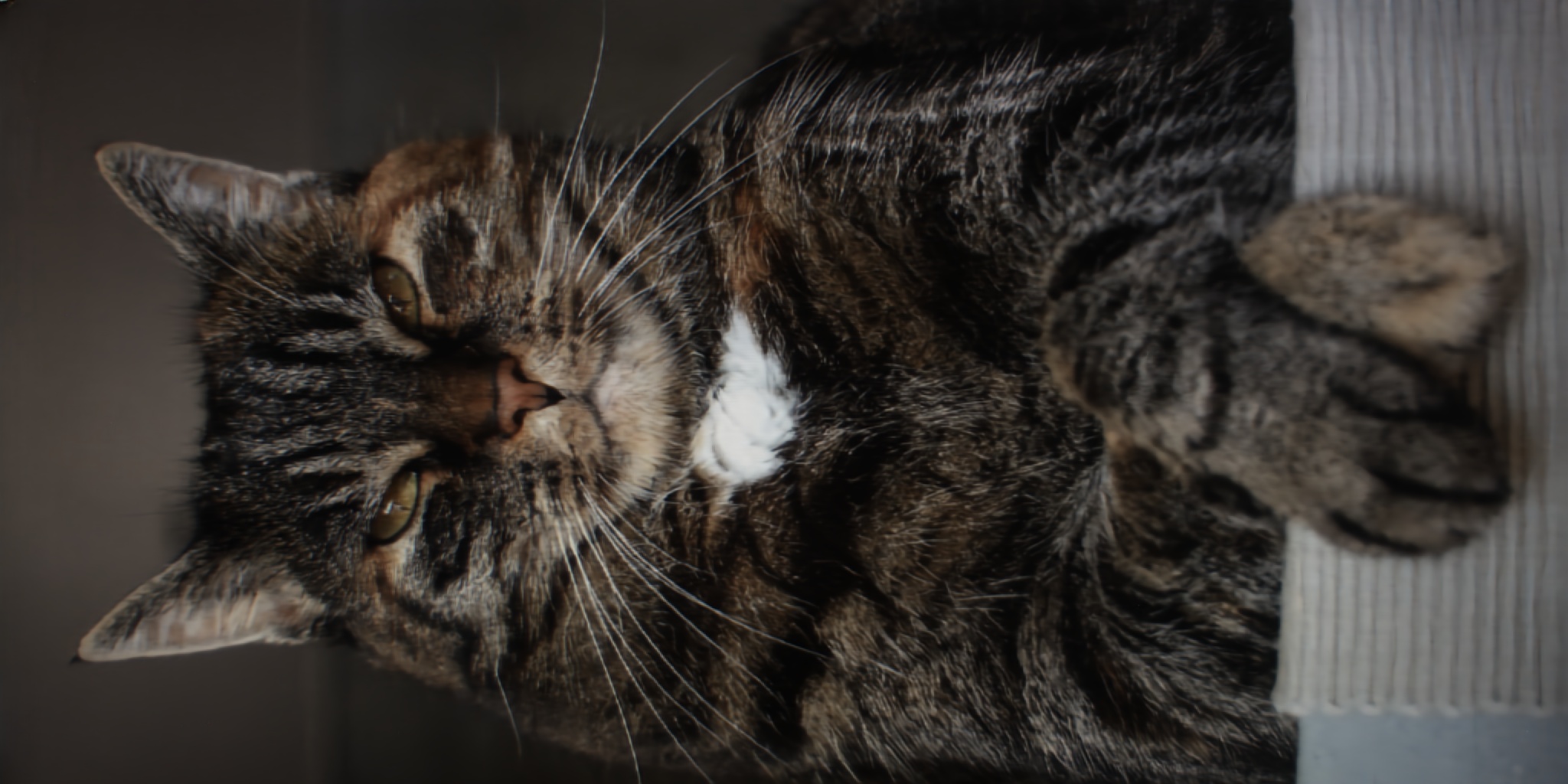}
		\fontsize{8}{12pt}\selectfont PSNR-$35.12 dB$/SSIM-$0.93$
		\vskip 2pt
        \includegraphics[scale=\scale]{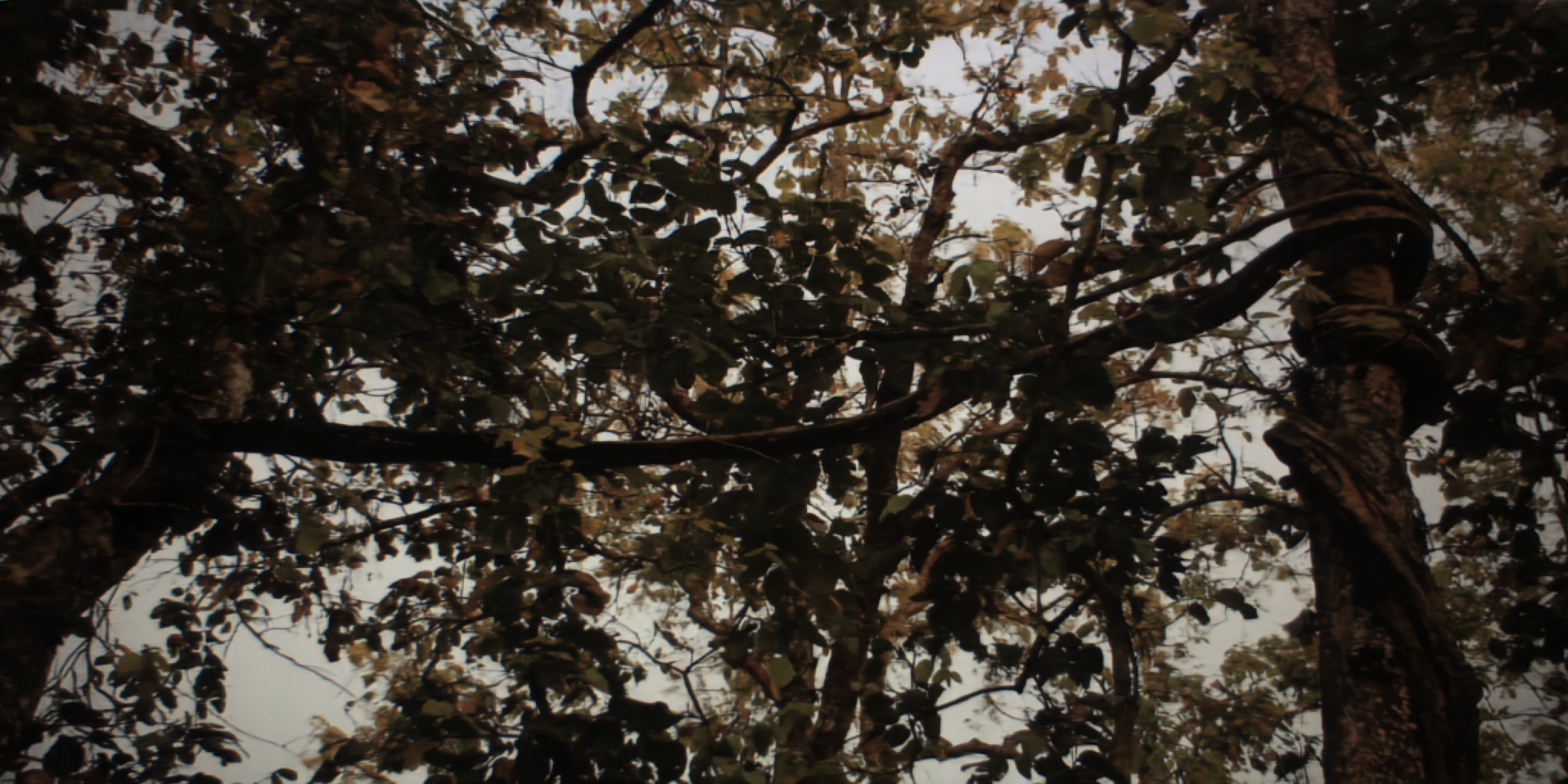}
        \fontsize{8}{12pt}\selectfont PSNR-$32.32 dB$/SSIM-$0.93$
        \vskip 2pt
        \includegraphics[scale=\scale]{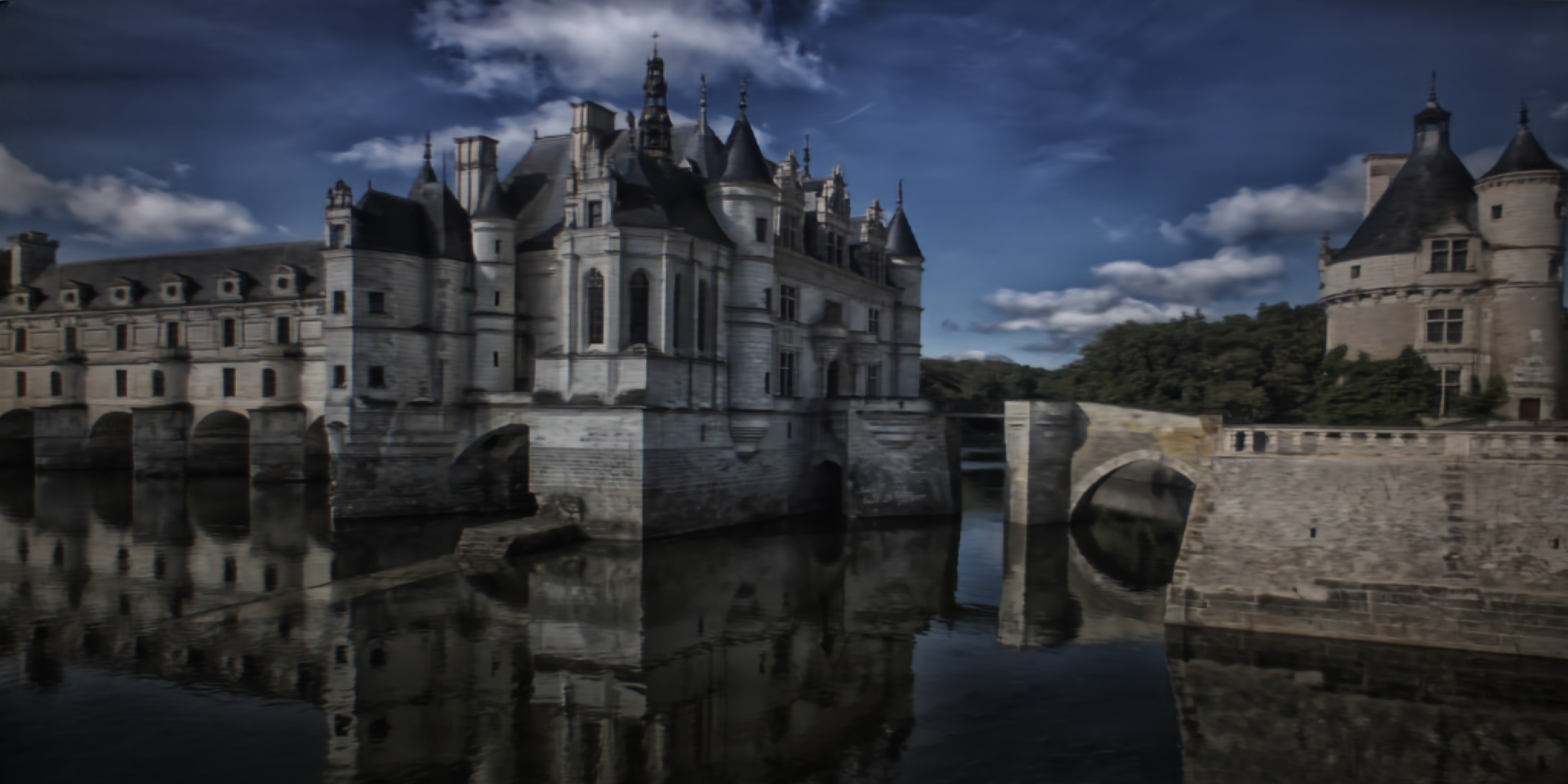}
        \fontsize{8}{12pt}\selectfont PSNR-$38.4 dB$/SSIM-$0.95$
        \vskip 2pt
        \includegraphics[scale=\scale]{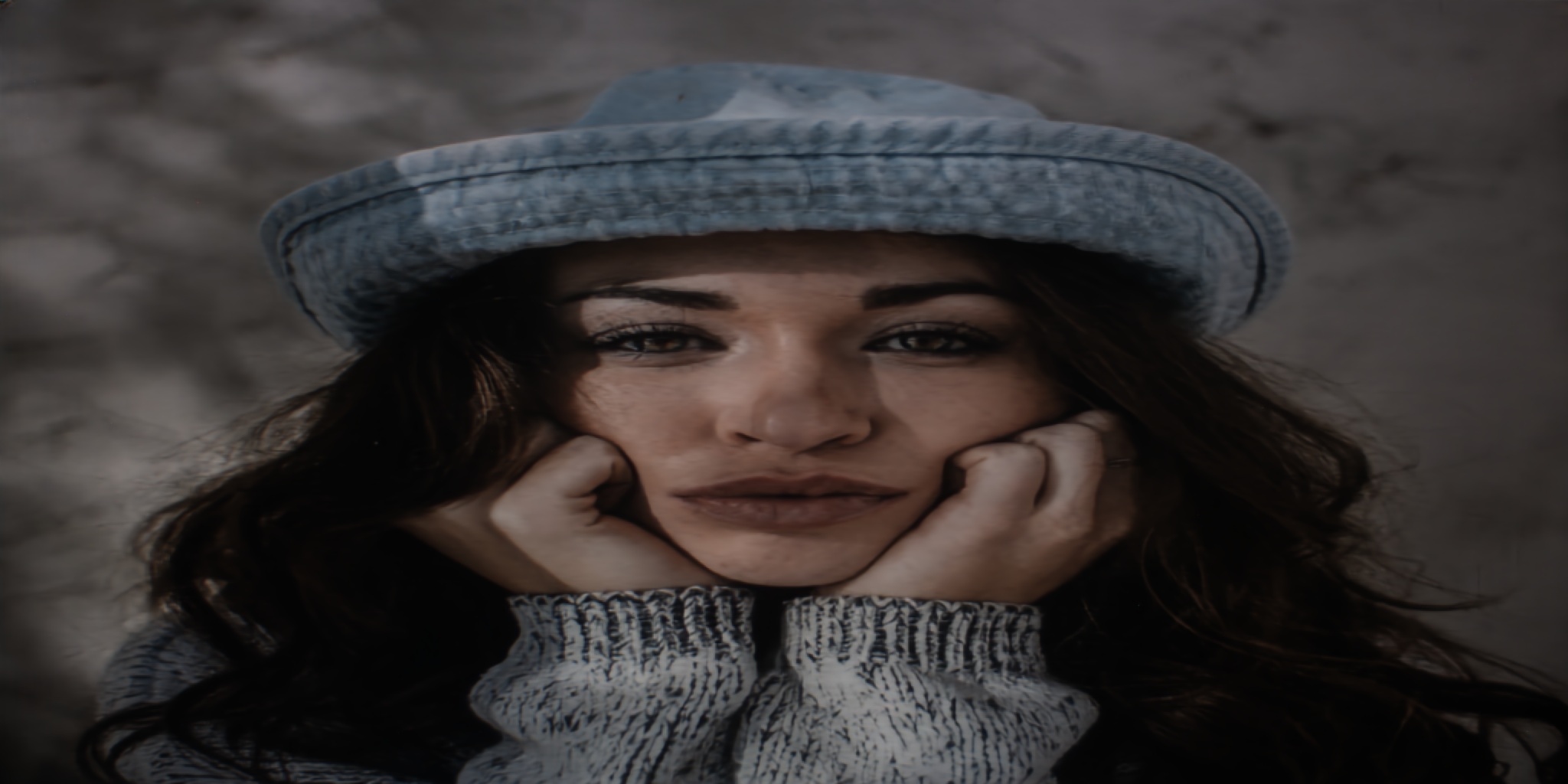}
        \fontsize{8}{12pt}\selectfont PSNR-$38.68 dB$/SSIM-$0.96$
        \vskip 2pt
		\fontsize{9}{12pt}\selectfont (b) Ours
		\end{center}
  \end{minipage}
  \begin{minipage}{\x\linewidth}
		\begin{center}
		\includegraphics[scale=\scale]{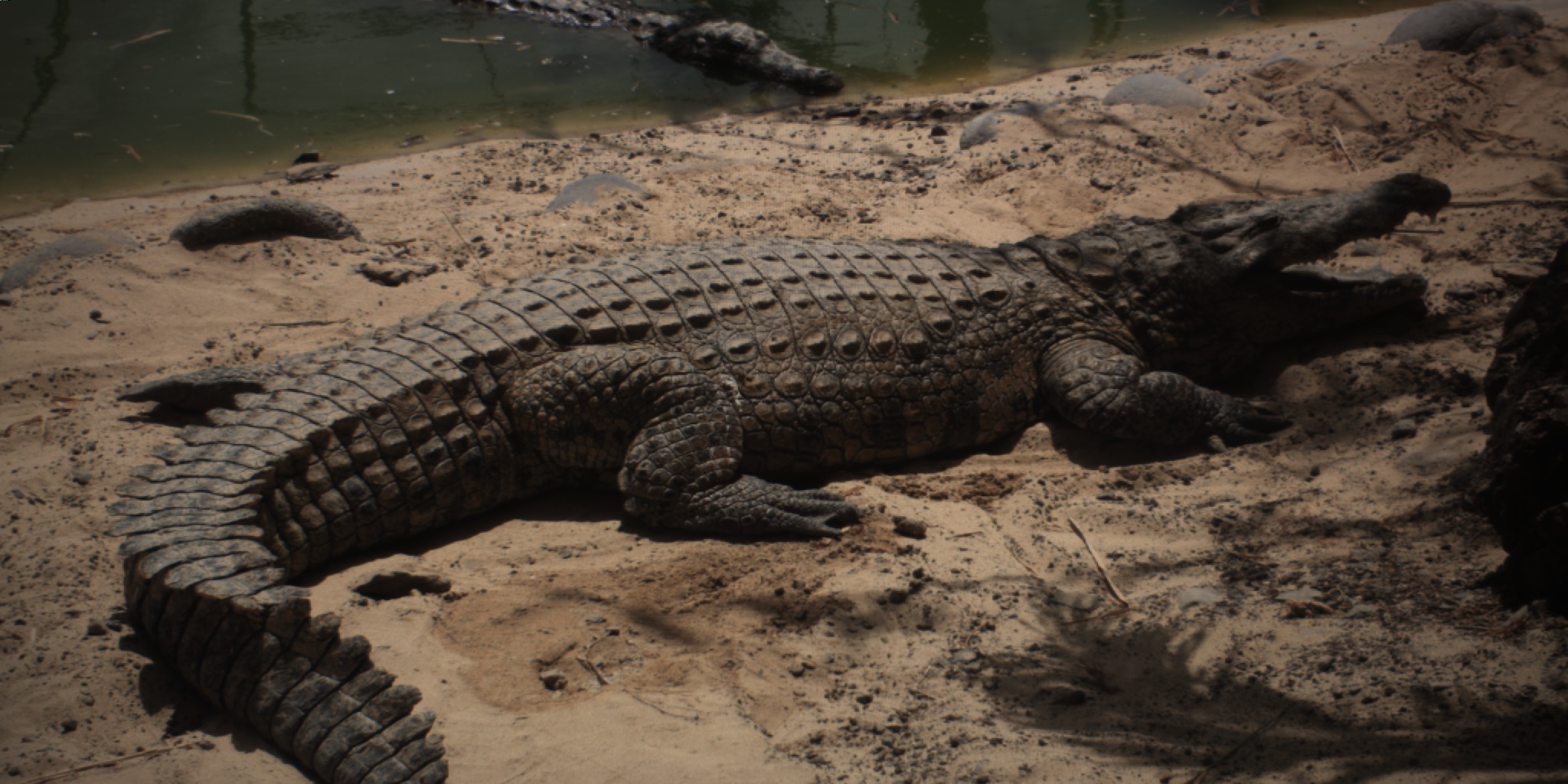}
		\fontsize{8}{12pt}\selectfont {\color{white} PSNR-$\infty  dB$/SSIM-$1$}
		\vskip 2pt
        \includegraphics[scale=\scale]{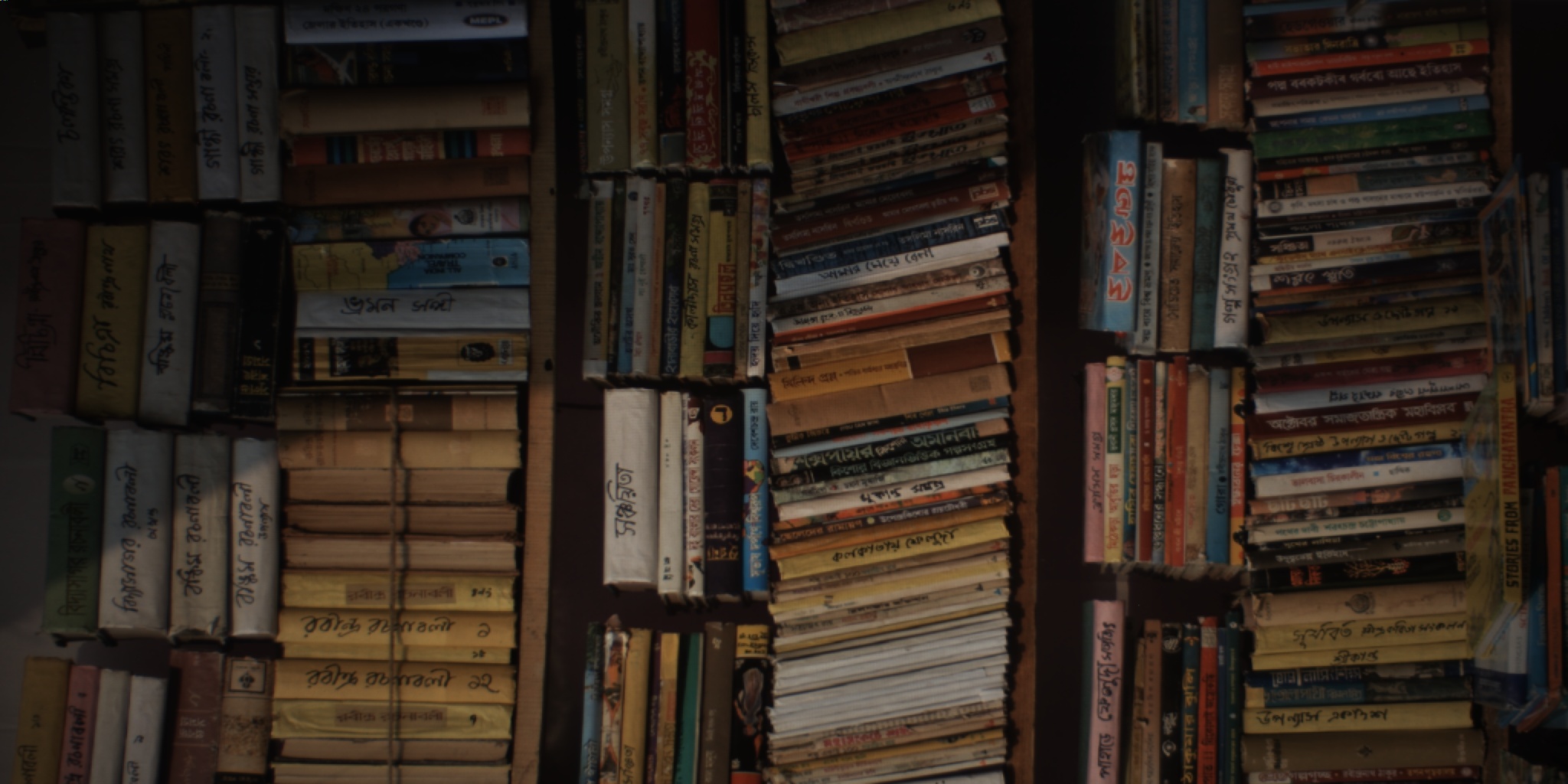}
        \fontsize{8}{12pt}\selectfont {\color{white} PSNR-$\infty  dB$/SSIM-$1$}
        \vskip 2pt
        \includegraphics[scale=\scale]{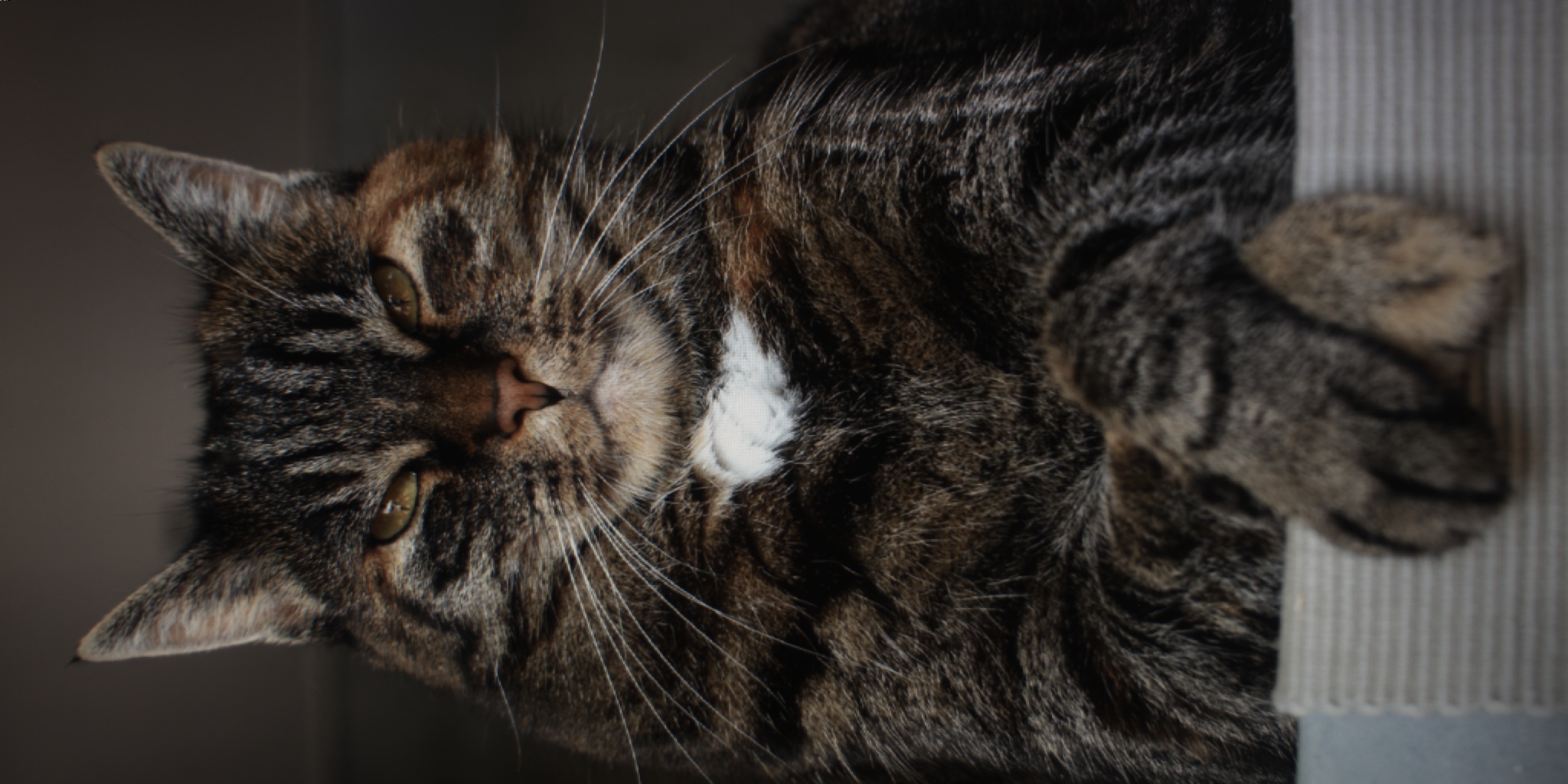}
		\fontsize{8}{12pt}\selectfont {\color{white} PSNR-$\infty  dB$/SSIM-$1$}
		\vskip 2pt
        \includegraphics[scale=\scale]{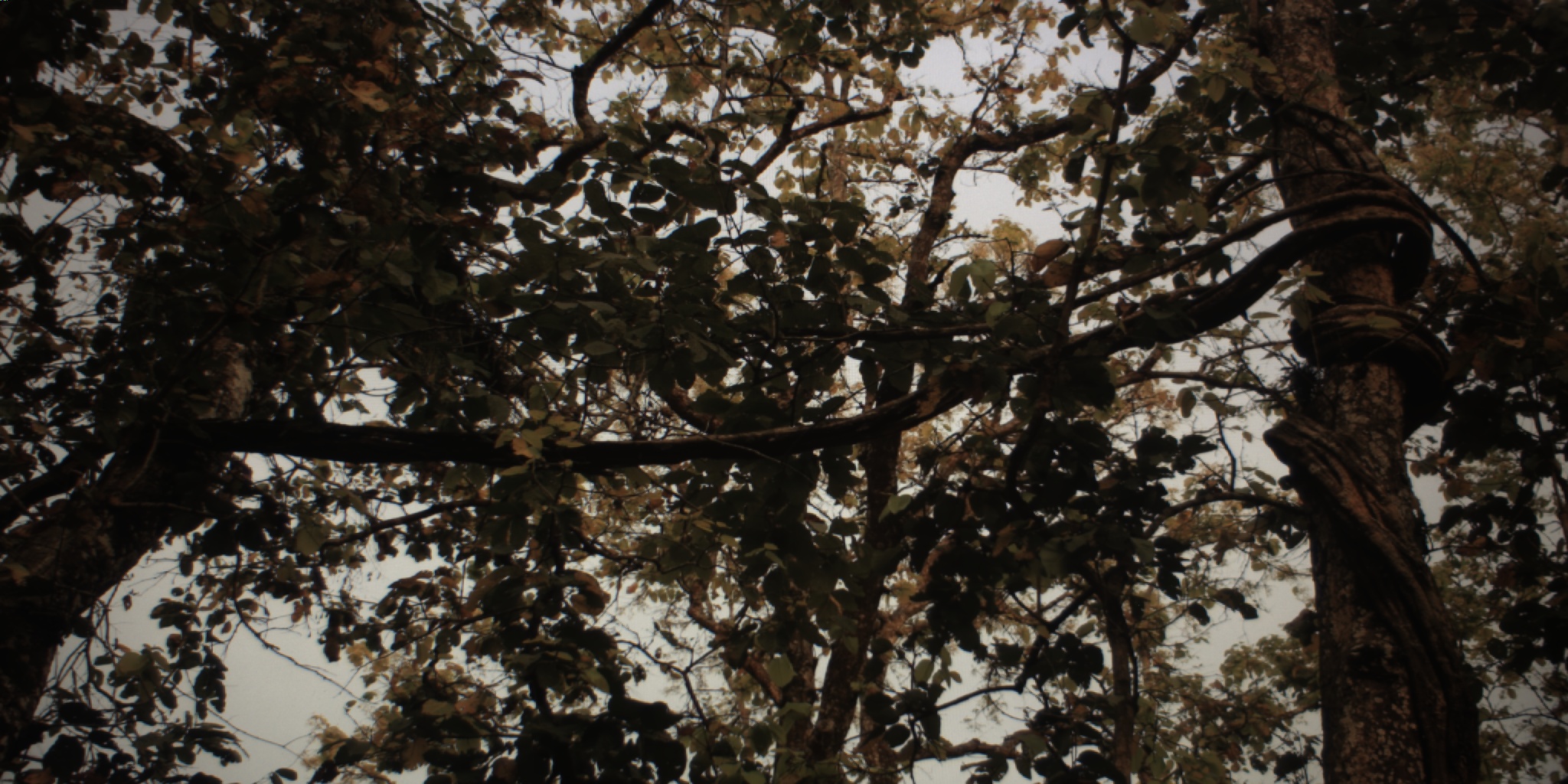}
        \fontsize{8}{12pt}\selectfont {\color{white} PSNR-$\infty  dB$/SSIM-$1$}
        \vskip 2pt
        \includegraphics[scale=\scale]{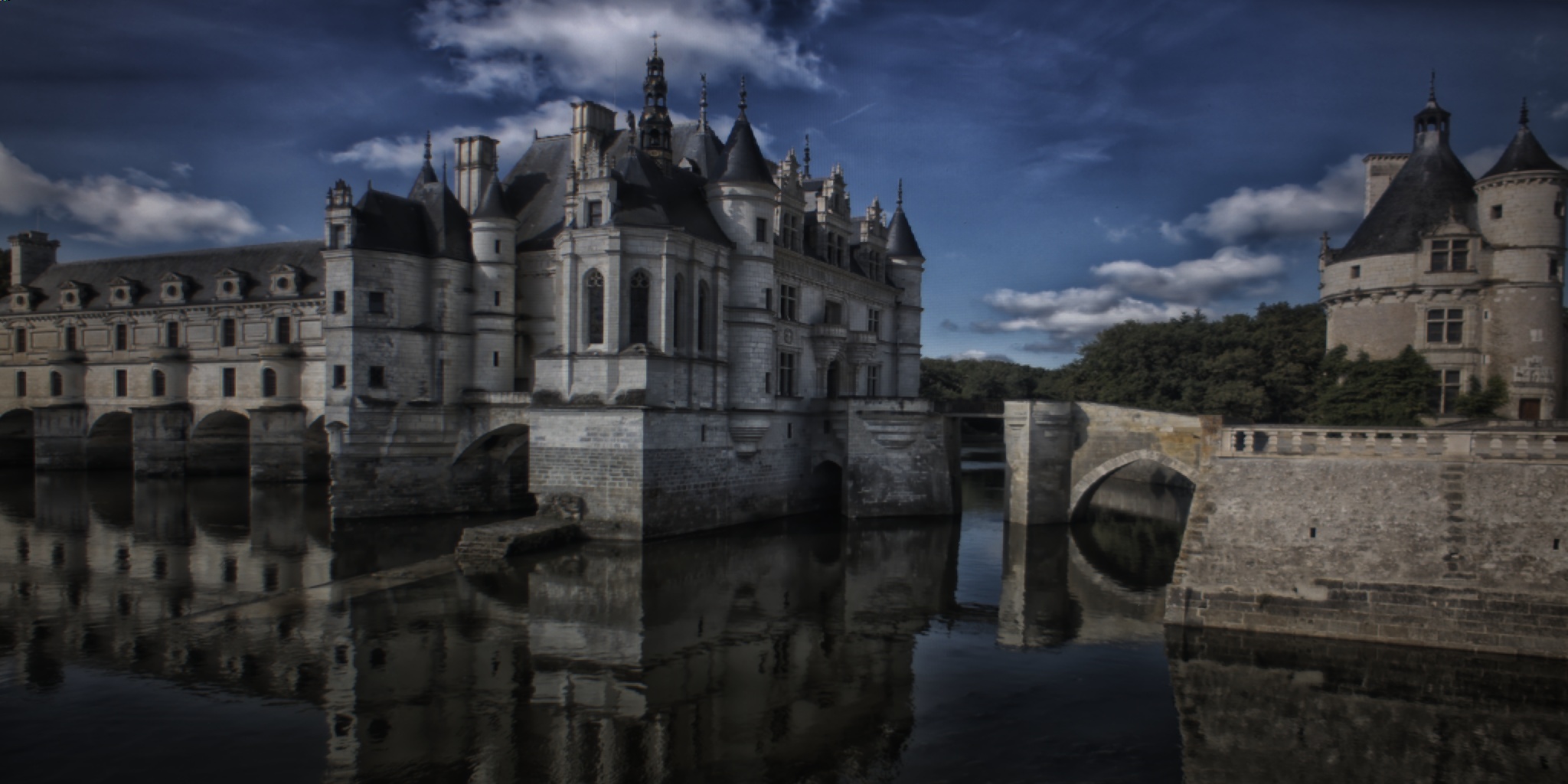}
        \fontsize{8}{12pt}\selectfont {\color{white} PSNR-$\infty  dB$/SSIM-$1$}
        \vskip 2pt
        \includegraphics[scale=\scale]{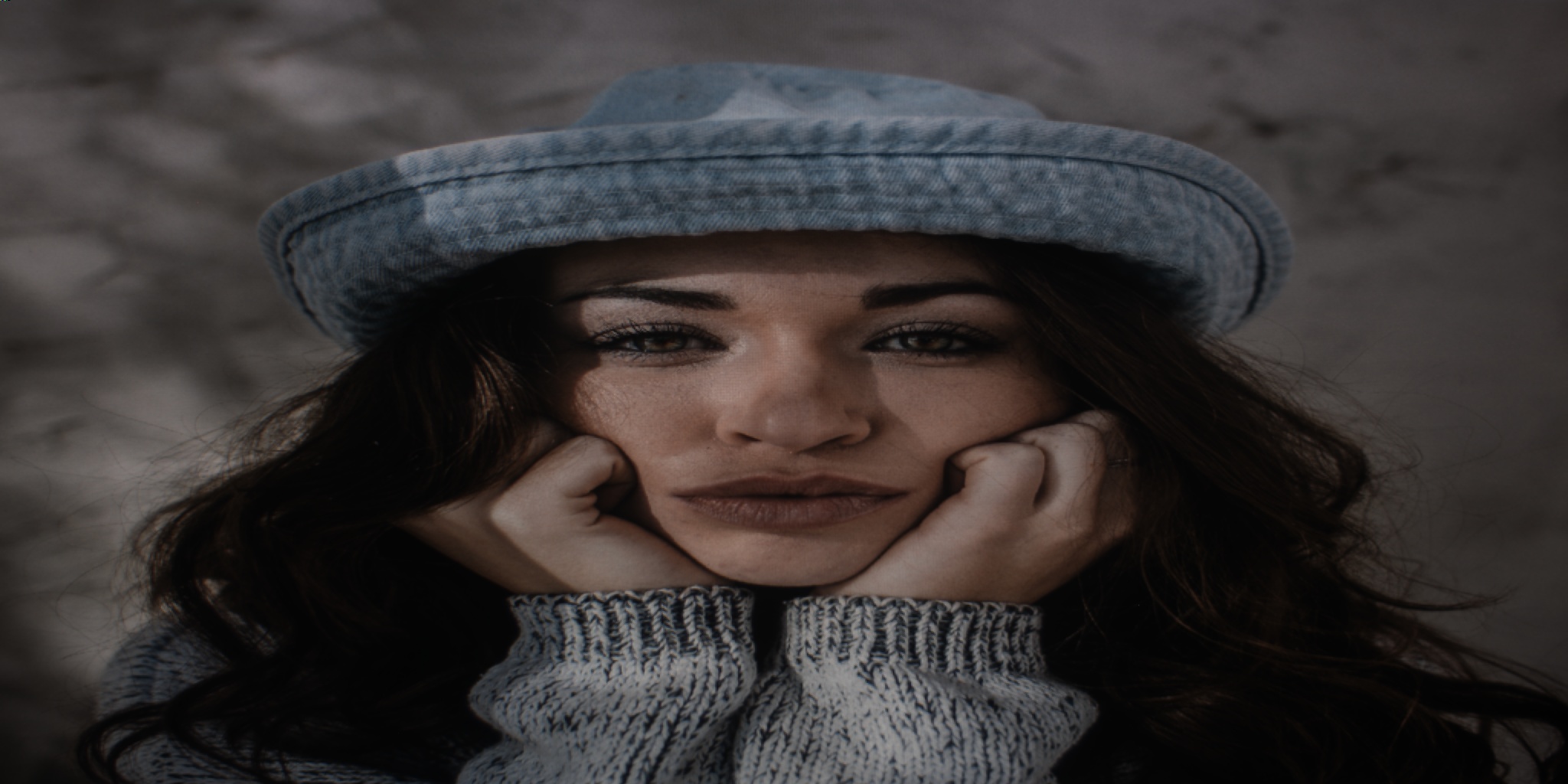}
        \fontsize{8}{12pt}\selectfont {\color{white} PSNR-$\infty  dB$/SSIM-$1$}
        \vskip 2pt
		\fontsize{9}{12pt}\selectfont (c) Ground truth
		\end{center}
  \end{minipage}
\caption{Sample results of the modified PDCRN with DDB for (a). T-OLED degraded images, (b). images restored using the network shows high fidelity with (c). the ground-truth captured without mounting any display.}
\label{fig:Toled sample results}
\end{figure}

The Fig. \ref{fig:poled sample results} and \ref{fig:Toled sample results} shows six sets of examples for P-OLED and T-OLED degraded image restoration. The proposed PDCRN was able to attain superior results in PSNR scores and from the visual comparison shows the high fidelity of the restored images with the ground truth. Similarly, the modified version of PDCRN with dual domain network also achieves high perceptual quality and fidelity scores for restored images degraded by T-OLED display. Two different networks have been proposed in this work to address the restoration of two types of under-display camera images. The PDCRN proposed for P-OLED image restoration was the winning entry in European conference on computer vision (ECCV) 2020 challenge on image restoration for Under-display Camera - Track 2 - P-OLED organized in conjunction with RLQ-TOD workshop. 
\par 
Table \ref{tab:table_poled} shows the performance comparison based on PSNR and SSIM of the proposed PDCRN with other competing entries in the challenge. It is evident from the table that the proposed PDCRN outperforms other methods by a considerable margin which proves its effectiveness. Similarly, PDCRN with dual domain network was a competing entry in Track -1 - T-OLED of the challenge. The modified PDCRN was able to achieve $4^{th}$ position in the final ranking and obtained comparable results only with a margin of $0.4$ dB with the challenge winners in terms of PSNR and SSIM as depicted in Table \ref{tab:table_toled}.

\begin{table}[]
\centering
\caption{Comparative study of the results from P-OLED track under UDC challenge in RLQ-TOD'20 workshop, ECCV-2020}
\label{tab:table_poled}
\resizebox{\textwidth}{!}{%
\begin{tabular}{|c|c|c|c|c|c|c|c|c|c|}
\hline
Teams & \textbf{Ours} & Team 1 & Team 2 & Team 3 & Team 4 & Team 5 & Team 6 & Team 7 & Team 8 \\ \hline
\textbf{PSNR} & \textbf{32.99(1)} & 32.29(2) & 31.39(3) & 30.89(4) & 29.38(5) & 28.61(6) & 26.6(7) & 25.72(8) & 25.46(9) \\ \hline
SSIM & \textbf{0.9578(1)} & 0.9509(2) & 0.9506(3) & 0.9469(5) & 0.9249(6) & 0.9489(4) & 0.9161(7) & 0.9027(8) & 0.9015(9) \\ \hline
\end{tabular}%
}
\end{table}
\begin{table}[]
\centering
\caption{Comparative study of the results from T-OLED track under UDC challenge in RLQ-TOD'20 workshop, ECCV-2020}
\label{tab:table_toled}
\resizebox{\textwidth}{!}{%
\begin{tabular}{|c|c|c|c|c|c|c|c|c|}
\hline
Teams & \textbf{Ours} & Team 1 & Team 2 & Team 3 & Team 4 & Team 5 & Team 6 & Team 7 \\ \hline
\textbf{PSNR} & \textbf{37.83 (4)} & 38.23 (1) & 38.18 (2) & 38.13 (3) & 36.91 (5) & 36.72 (6) & 34.35 (7) & 33.78 (8) \\ \hline
SSIM & \textbf{0.9783(4)} & 0.9803 (1) & 0.9796 (2) & 0.9797 (3) & 0.9734(6) & 0.9776 (5) & 0.9645 (7) & 0.9324 (8) \\ \hline
\end{tabular}%
}
\end{table}
\pagebreak
\section{Conclusions}
\label{sec:conclusions}
In this paper, we proposed two different fully convolutional networks for the restoration of images degraded due to under-display imaging. Pyramidal Dilated Convolutional RestoreNet proposed for pentile-organic LED images has obtained state-of-the-art restoration performance in terms of standard evaluation metrics PSNR and SSIM. The network proposed for transparent-organic LED based imaging used a dual domain approach with implicit DCT and obtained considerable performance when compared with state-of-the-art methodologies. An intensive study may be conducted to devise a methodology that can generalize both P-OLED and T-OLED degradations equally well.

\section{Acknowledgements}
\label{sec:Acknowledgements}
 We gratefully acknowledge the support of NVIDIA PSG Cluster and Trivandrum Engineering Science and Technology Research Park (TrEST) with the donation of the Tesla V100 GPUs used for this research.

\clearpage
%
%
\bibliographystyle{splncs04}
\bibliography{main}
\end{document}